\documentclass[10pt,twocolumn,letterpaper]{article}

\usepackage[pagenumbers]{cvpr} 

\usepackage{graphicx}
\usepackage{amsmath}
\usepackage{amssymb}
\usepackage{booktabs}
\usepackage{algorithm}
\usepackage{algpseudocode}
\usepackage[table,xcdraw]{xcolor}
\usepackage{tabularx}
\usepackage{makecell}
\usepackage{multirow}
\usepackage[pagebackref,breaklinks,colorlinks]{hyperref}

\usepackage[capitalize]{cleveref}
\crefname{section}{Sec.}{Secs.}
\Crefname{section}{Section}{Sections}
\Crefname{table}{Table}{Tables}
\crefname{table}{Tab.}{Tabs.}

\newcommand{\mth}[1]{\multirowcell{-2}{#1}}
 
\def\cvprPaperID{6656} 
\def\confName{CVPR}
\def\confYear{2023}

\begin{document}

\title{Join the High Accuracy Club on ImageNet with A Binary Neural Network Ticket}

\author{Nianhui Guo\\
Hasso Plattner Institute \\
Germany\\
{\tt\small nianhui.guo@hpi.de}
\and
Joseph Bethge \\
Hasso Plattner Institute \\
Germany\\
{\tt\small joseph.bethge@hpi.de}
\and
Christoph Meinel \\
Hasso Plattner Institute \\
Germany\\
{\tt\small christoph.meinel@hpi.de}
\and
Haojin Yang \\
Hasso Plattner Institute \\
Germany\\
{\tt\small haojin.yang@hpi.de}
}
\maketitle

\begin{abstract}
\label{sec:abstract}
Binary neural networks are the extreme case of network quantization, which has long been thought of as a potential edge machine learning solution. 
However, the significant accuracy gap to the full-precision counterparts restricts their creative potential for mobile applications. 
In this work, we revisit the potential of binary neural networks and focus on a compelling but unanswered problem:
how can a binary neural network achieve the crucial accuracy level (e.g., 80\%) on ILSVRC-2012 ImageNet?
We achieve this goal by enhancing the optimization process from three complementary perspectives: 
(1) We design a novel binary architecture BNext based on a comprehensive study of binary architectures and their optimization process.
(2) We propose a novel knowledge-distillation technique to alleviate the counter-intuitive overfitting problem observed when attempting to train extremely accurate binary models.
(3) We analyze the data augmentation pipeline for binary networks and modernize it with up-to-date techniques from full-precision models.
The evaluation results on ImageNet show that BNext, for the first time, pushes the binary model accuracy boundary to \textbf{80.57\%} and significantly outperforms all the existing binary networks.
%
Code and trained models are available at:  \url{https://github.com/hpi-xnor/BNext.git}.
\end{abstract}
\section{Introduction}
\label{sec:introduction}
Deep neural networks (DNNs) have made remarkable progress in almost all areas of AI research in recent years. 
Despite their powerful ability, the development of modern neural network architectures is usually accompanied by an increase in computational budget, memory usage, and energy consumption \cite{simonyan2014very,szegedy2015going,he2016deep,huang2017densely,tan2019efficientnet,dosovitskiy2020image,wang2020deep,liu2022convnet}. 
On the other hand, common edge-computing platforms such as cell phones, mini-robots, AR glasses, and autopilot systems can only provide limited computing power and battery life.  
Consequently, there is an immense technical gap between the evolutionary trend of modern neural networks and edge applications.
Only co-designed optimization from both sides can maximize the potential of deep neural networks on the edge.

\begin{figure}[]
\captionsetup[subfigure]{justification=centering, font=tiny}
\begin{center}
\hspace*{-10pt} 
\begin{subfigure}[t]{.36\textwidth}
  \centering
  \includegraphics[width=\linewidth, scale=0.5]{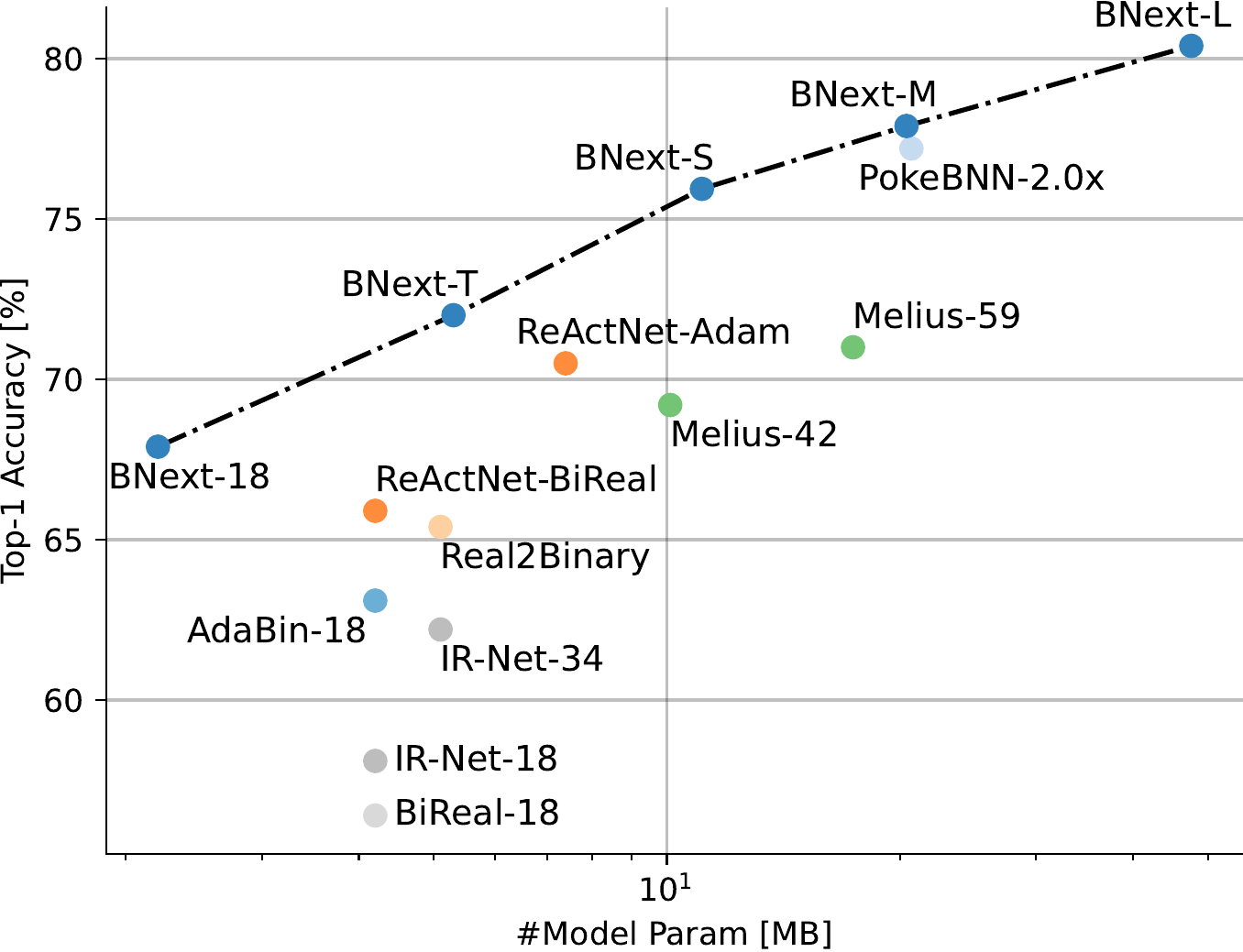}
  \caption{SOTA Comparison on ImageNet}
  \label{fig:sota comparison}
\end{subfigure}
\hspace{-1pt}
\vspace{-10pt}
\begin{subfigure}[t]{.12\textwidth}
    \centering
    \includegraphics[width=17mm, height=52mm]{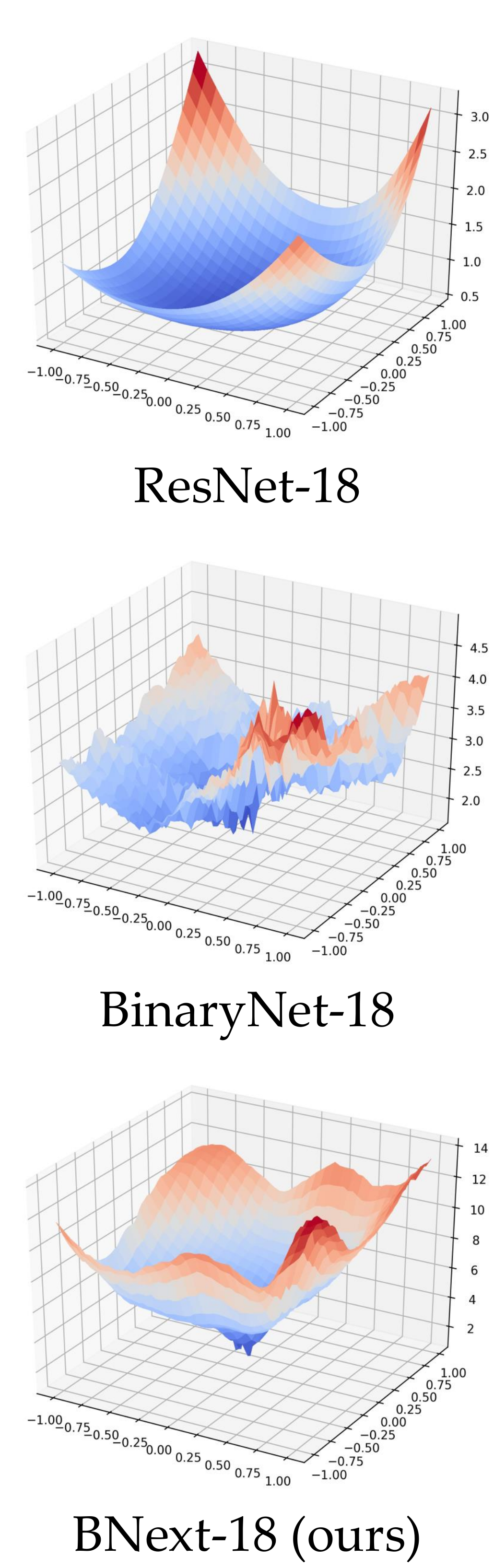}
    \centering
    \caption{Loss Landscape}
    \label{fig:loss landscape motivation}
\end{subfigure}
\end{center}
\vskip -2pt 
\caption{(a) Comparison with SOTA binary neural networks on ILSVRC-2012 ImageNet in terms of accuracy and model param. 
Google's PokeBNN relies on a huge batch size of 8192, and ours is 512, in line with most previous works.
(b) 3D loss landscape visualization \cite{li2018visualizing} of full-precision ResNet-18 \cite{he2016deep}, BinaryNet-18 \cite{hubara2016binarized} and our BNext-18.
The smoothness of BNext-18 is particularly close to the full-precision counterpart.}
\label{fig:teaser}
\vskip -8pt 
\end{figure}

Motivated by this, researchers have developed various techniques to compress and accelerate modern neural networks (e.g., Network Pruning \cite{han2015deep}, Knowledge Distillation\cite{hinton2015distilling}, Compact Architecture Design \cite{howard2017mobilenets}, low-bit Quantization \cite{Jacob_2018_CVPR,banner2019post}). 
Among them, Binary Neural Networks (BNNs) have received the attention of researchers owing to their vast potential on edge devices \cite{courbariaux2016binarized,rastegari2016xnor, liu2018birealnet,real2binICLR20,liu2020reactnet, zhang2021pokebnn}.
By restricting the value range to 1-bit (e.g., ${\pm1}$), the binary quantization technique can compress memory requirements by ${32\times}$ and achieve a theoretical speedup of ${58\times}$ on the CPU by replacing the floating-point dot products with XNOR and bit-counting operations \cite{courbariaux2016binarized,rastegari2016xnor}. 
Despite the apparent efficiency advantages, binary neural networks have long suffered from optimization difficulties and accuracy degradation problems. 
Previous works have made great efforts to narrow the accuracy gap to full-precision models such as ResNet-18 and ResNet-50 \cite{courbariaux2016binarized,rastegari2016xnor,liu2018birealnet,real2binICLR20,liu2020reactnet,zhang2021pokebnn}. 
However, the latest BNNs are still far behind the latest full-precision counterparts with \raisebox{-0.55ex}{\~{}}10\% lower accuracy \cite{dosovitskiy2020image, liu2022convnet, liu2021swin}, due to their extreme degradation of representation capacity (${3.4 \times 10^{38}}$ lower) \cite{tu2022adabin} and increased optimization difficulty. 
To solve the problem of non-differentiability in backpropagation, BNN training relies on gradient estimation techniques, e.g., the Straight-Through Estimator \cite{courbariaux2016binarized}. 
BNNs also have a much coarser loss landscape where the optimization roadmap to the global minimum is more rugged than for full-precision networks (see Fig.~\ref{fig:loss landscape motivation}).
%
The aforementioned problems complicate the optimization of BNNs, which then require specific architecture designs and optimization methodology for highly accurate results.
%
Therefore, the question of \emph{how a binary neural network can achieve the crucial accuracy level (e.g., 80\%) on ILSVRC-2012 ImageNet} is still an open question.
In this paper, we build BNext, the first BNN with 80\%+ accuracy on the ILSVRC-2012 ImageNet dataset. 
To enhance the representative capacity, we first construct a novel binary processing unit with an adaptive information re-coupling structure called Info-RCP. 
Previous works \cite{rastegari2016xnor, real2binICLR20, zhang2021pokebnn} utilize channel-wise scaling to adjust the distribution after each binary convolution. 
Although this alleviates the information loss caused by binarization, these traditional Real2Binary-style designs \cite{real2binICLR20} ignore the semantic gaps before and after binary convolution, which causes sub-optimal information coupling. 
Instead, we first reduce this distribution gap by scaling and shifting the binary convolution outputs with an extra BatchNorm and a PReLU layer. 
Then, we reshape the output distribution of this unit using a lightweight attention branch, which combines the information before and after this binary unit as the inputs. 
Subsequently, the basic BNext block is constructed by stacking multiple Info-RCP modules.
Based on our analysis of classical binary architectures and motivated by the regularized architecture design of the recent vision transformer \cite{han2021demystifying, liu2021swin}, we propose a novel element-wise attention (ELM-Attention) module to enhance the binary basic block further. 
More specifically, each basic block utilizes an element-wise bypass using multiplication to dynamically calibrate the output of the first Info-RCP module in the basic block. 
Finally, we build the BNext family with different model sizes by stacking the basic block using different stage-design strategies. 
As shown in Fig.~\ref{fig:loss landscape motivation}, the loss landscape of BNext-18 is much smoother and more complete compared to a traditional binary ResNet-18 and is pretty close to a full-precision ResNet-18. 

The optimization pipeline design plays a crucial role in high-accuracy BNN optimization.  
Compared to 32-bit DNNs, BNNs own a coarser loss landscape, which adds more obstacles for optimizing towards the global minimum.
It thus requires more fine-grained and flexible gradient information to escape from sub-optimal convergence.
To this end, Knowledge Distillation (KD) is a commonly used technique in previous works \cite{real2binICLR20,liu2020reactnet,liu2021adam}.
However, when we attempt to push the accuracy level of BNNs from the current 70\% to 80\% level, we observe that BNNs are more prone to counter-intuitive overfitting using the standard KD techniques.
Therefore, improving the teacher model selection and KD pipeline design is essential.
We thus propose a new metric, \emph{Knowledge Complexity}, as a simple yet effective indicator for teacher model selection.
%
%
We further use the selected teachers (e.g., \cite{tan2019efficientnet,liu2022convnet}), including a principal teacher model and an assistant teacher group.
We adaptively pick the best suitable assistant teacher at different training stages to improve the student's accuracy and avoid overfitting.
The assistant teachers serve as regularizers for the knowledge from high-confidence prediction. 
Furthermore, modern DNNs \cite{dosovitskiy2020image, liu2022convnet, liu2021swin} usually rely on well-designed data augmentation pipelines. 
%
%
To the best of our knowledge, this is the first work that thoroughly verifies recently proposed data augmentation techniques on BNNs and provides practical guidance based on extensive experiments. 
Overall, as shown in Fig.~\ref{fig:sota comparison}, BNext outperforms all previous BNNs by a large margin and is the first binary neural network to reach the 80\% accuracy level on ImageNet.

Our main contributions can be summarized as follows:
\begin{enumerate}
    \item 
    \emph{BNext}, a novel binary architecture with a smoother loss landscape than existing popular designs which makes it easier to optimize.
    
    \item
    The \emph{Diversified Consecutive Knowledge Distillation} technique, which alleviates the counter-intuitive overfitting problem in highly accurate BNNs. 
    
    \item 
    A modern training pipeline for BNNs based on a fair and comprehensive verification of up-to-date data augmentation techniques.

    \item 
    The first BNN to achieve $>$\textbf{80\%} top-1 accuracy on ImageNet, which reshapes our vision of BNNs' potential.
\end{enumerate}



\section{Related Work}
\label{sec:related work}

\textbf{BNN Optimization.} Binary neural networks are difficult to optimize.
Compared to their 32-bit counterparts, BNNs are almost universally non-differentiable.
BinaryNet \cite{courbariaux2016binarized} solved this challenge and proved the feasibility of optimizing BNNs by using the straight-through-estimation (STE) technique.
Later works (e.g. \cite{liu2018birealnet, lin2020rotated, wang2021gradient, guo2021boolnet}) have attempted to improve the optimization using different variants of STE for approximating the gradient of sign function.
\cite{NEURIPS2019_9ca8c9b0} revisited the functional role of latent weights in BNNs and proposed a specialized optimizer BOP to flip the binary states.
Real2BinaryNet \cite{real2binICLR20} applies knowledge distillation in three stages to optimize BNNs.
Similar works (\cite{liu2020reactnet, zhang2021pokebnn}) simplified the KD pipeline to a two-stage process.
Moreover, Liu \etal \cite{liu2020reactnet} explore how training strategies (e.g., optimizers, weight decay) to aid BNN optimization.
In this paper, we provide the first in-depth analysis of the relationship between BNN optimization and the corresponding architecture design.
The analysis helps us to understand why BNNs are hard to optimize and how to construct a highly optimization-friendly architecture. 

\textbf{Knowledge Distillation.} Knowledge distillation \cite{hinton2015distilling} is used to ``distill'' dark knowledge from a pre-trained teacher model to a  student model.
In more recent works, researchers have turned their attention to a deeper understanding of this technique.
For example, it was found that there is a surprisingly large discrepancy between the predictive distributions of the teacher and the student, and a good fit for the teacher paradoxically does not generalize well for the student \cite{stanton2021does}.
Beyer \etal \cite{beyer2022knowledge} show that certain implicit design decisions drastically affect the effectiveness of distillation. 
Park \etal \cite{park2021learning} improves the distillation process by building student-friendly teachers and implicitly reveal the presence of a knowledge mismatch between teacher and student. 
Meanwhile, modern BNN optimization \cite{real2binICLR20,liu2020reactnet,zhang2021pokebnn} heavily relies on knowledge distillation techniques \cite{hinton2015distilling, shen2020meal}. 
The soft label from a pre-trained teacher provides a more fine-grained supervision signal than a one-hot label. 
Therefore, the BNN research community is encouraged to explore the latest advances in KD technologies and incorporate new insights into BNN research.

\textbf{Modern Neural Network Optimization.} 
The success of modern deep neural networks \cite{tan2019efficientnet,dosovitskiy2020image, liu2021swin,liu2022convnet} for image classification relies not only on architecture design ideas, such as the attention mechanism but also on a bag of up-to-date optimization tricks. 
Various data augmentation strategies such as MixUp \cite{zhang2017mixup}, CutMix \cite{yun2019cutmix}, RandAugmentation \cite{cubuk2020randaugment} and Augmentation Repetition \cite{hoffer2020augment} are evaluated to be beneficial for the generalization of regular 32-bit networks \cite{ridnik2022solving}. 
However, for highly accurate BNNs it is still an open question whether regular strategies can be applied or new BNNs-specific augmentation strategies should be used. 
This work gives the first fair and comprehensive empirical study on modern optimization tricks for BNNs to check their feasibility in large-scale image classification tasks. 
The detailed ablation study demonstrates the necessity of revisiting BNN-specific augmentation strategies and reveals possible solutions.

\begin{figure*}[]
\captionsetup[subfigure]{justification=centering, font=tiny}
\begin{center}
\begin{subfigure}[t]{.13\textwidth}
    \centering
    \includegraphics[width=22mm, height=22mm]{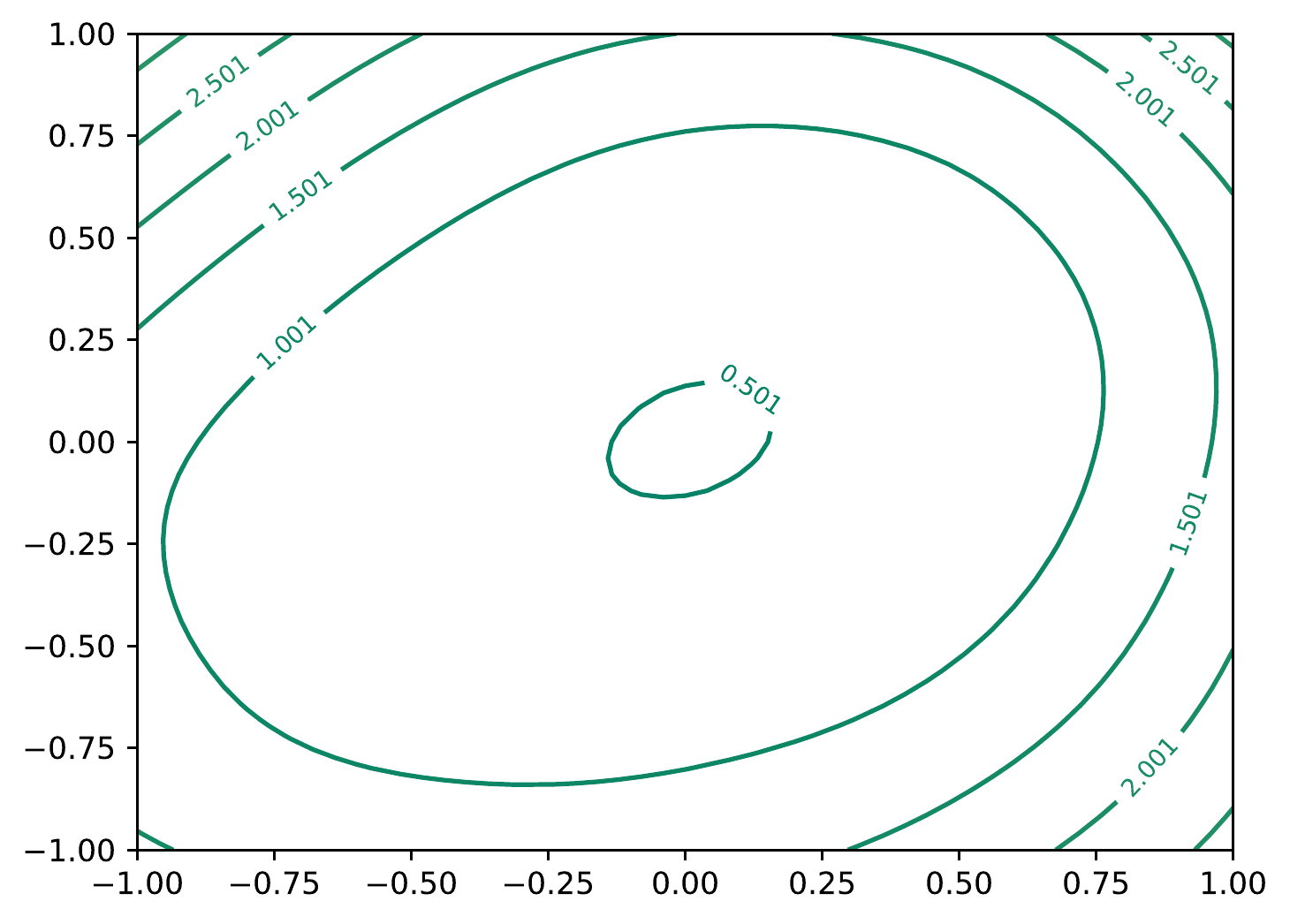}
    \centering
    \caption{ResNet-18}
    \label{fig:loss landscape resnet18}
\end{subfigure}
\hfill
\begin{subfigure}[t]{.13\textwidth}
    \centering
    \includegraphics[width=22mm, height=22mm]{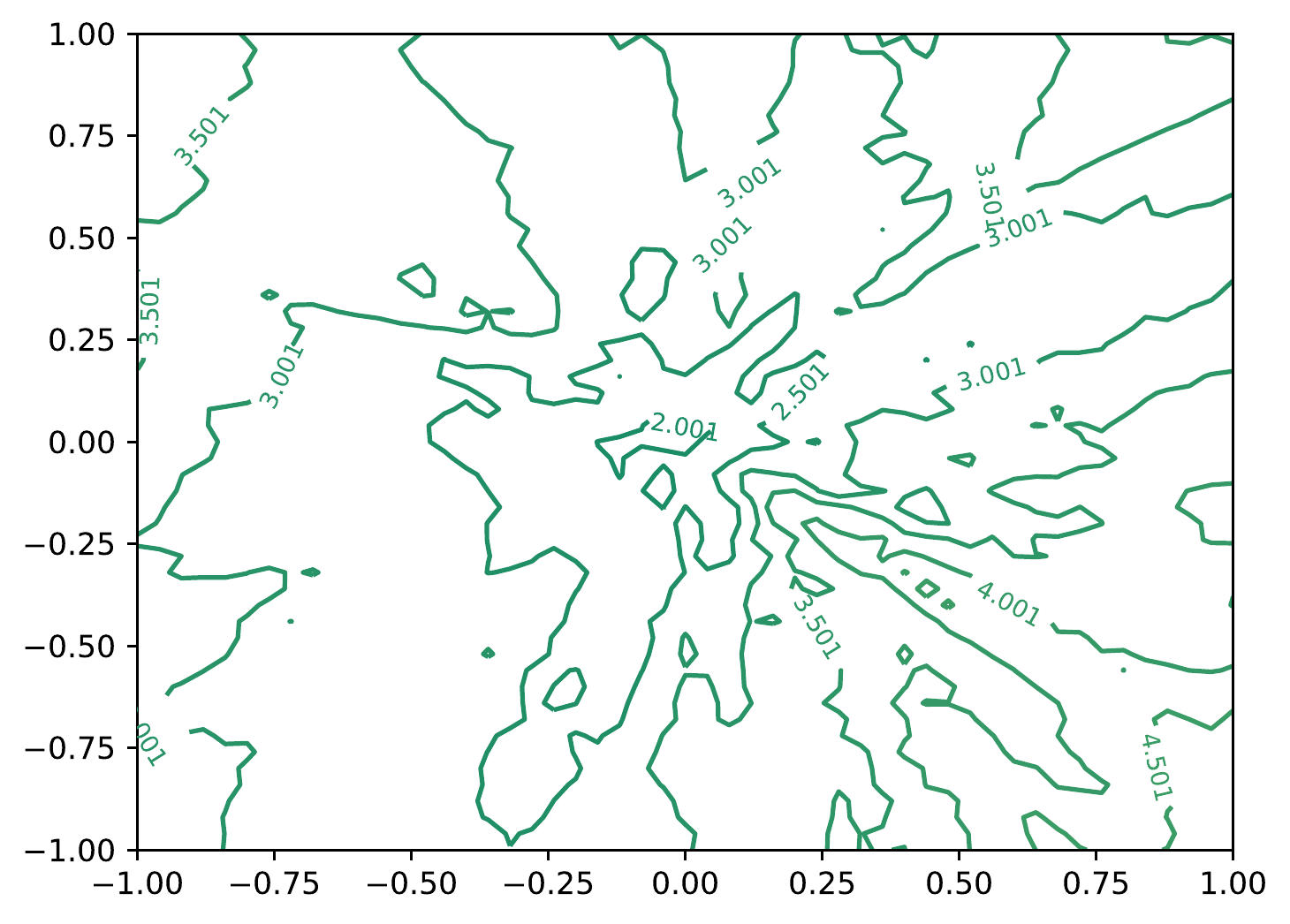}
    \centering
    \caption{BinaryNet-18}
    \label{fig:loss landscape binarynet-18}
\end{subfigure}
\hfill
\begin{subfigure}[t]{.13\textwidth}
    \centering
    \includegraphics[width=22mm, height=22mm]{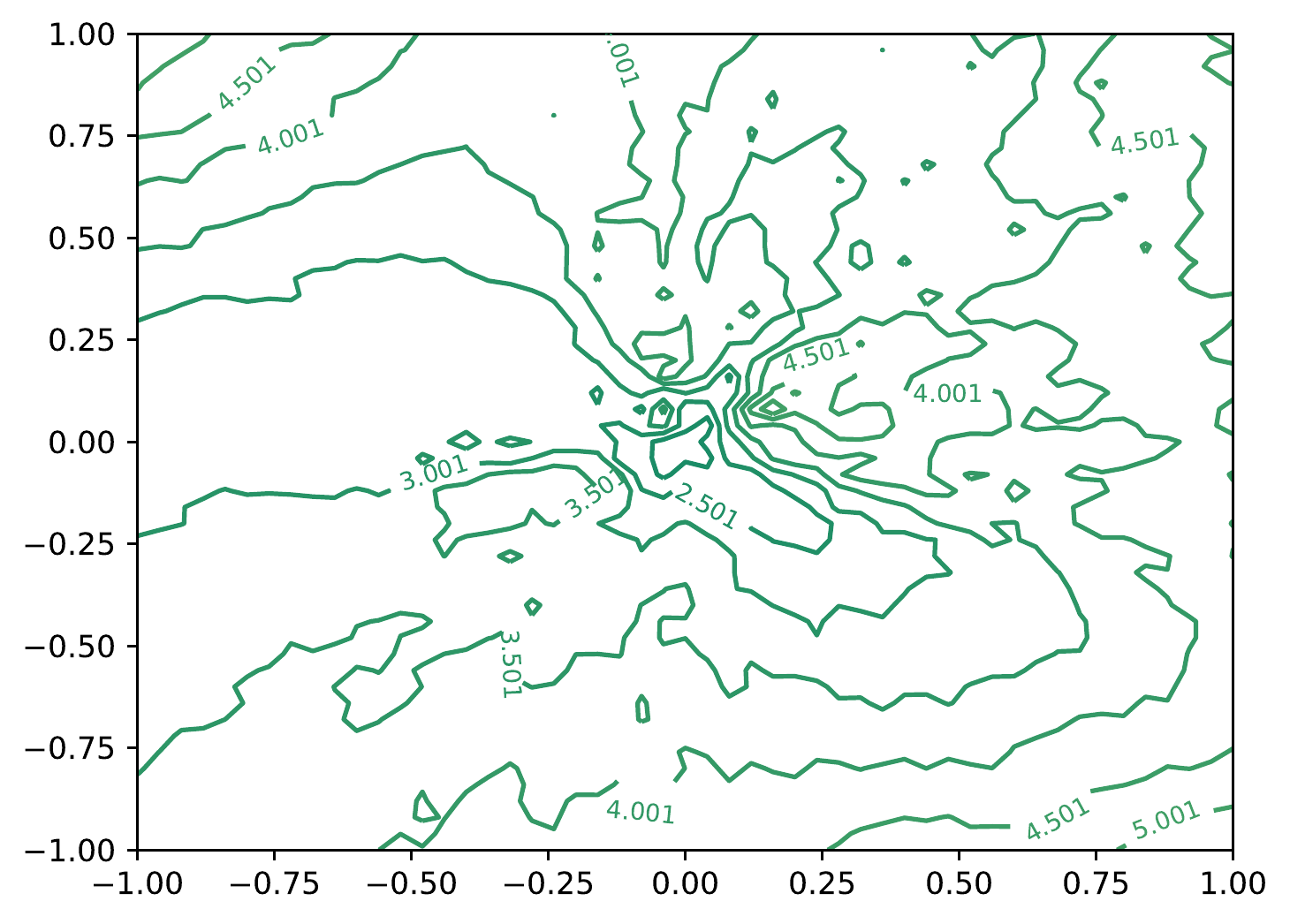}
    \centering
    \caption{BiRealNet-18}
    \label{fig:loss landscape birealnet-18}
\end{subfigure}
\hfill
\begin{subfigure}[t]{.13\textwidth}
    \centering
    \includegraphics[width=22mm, height=22mm]{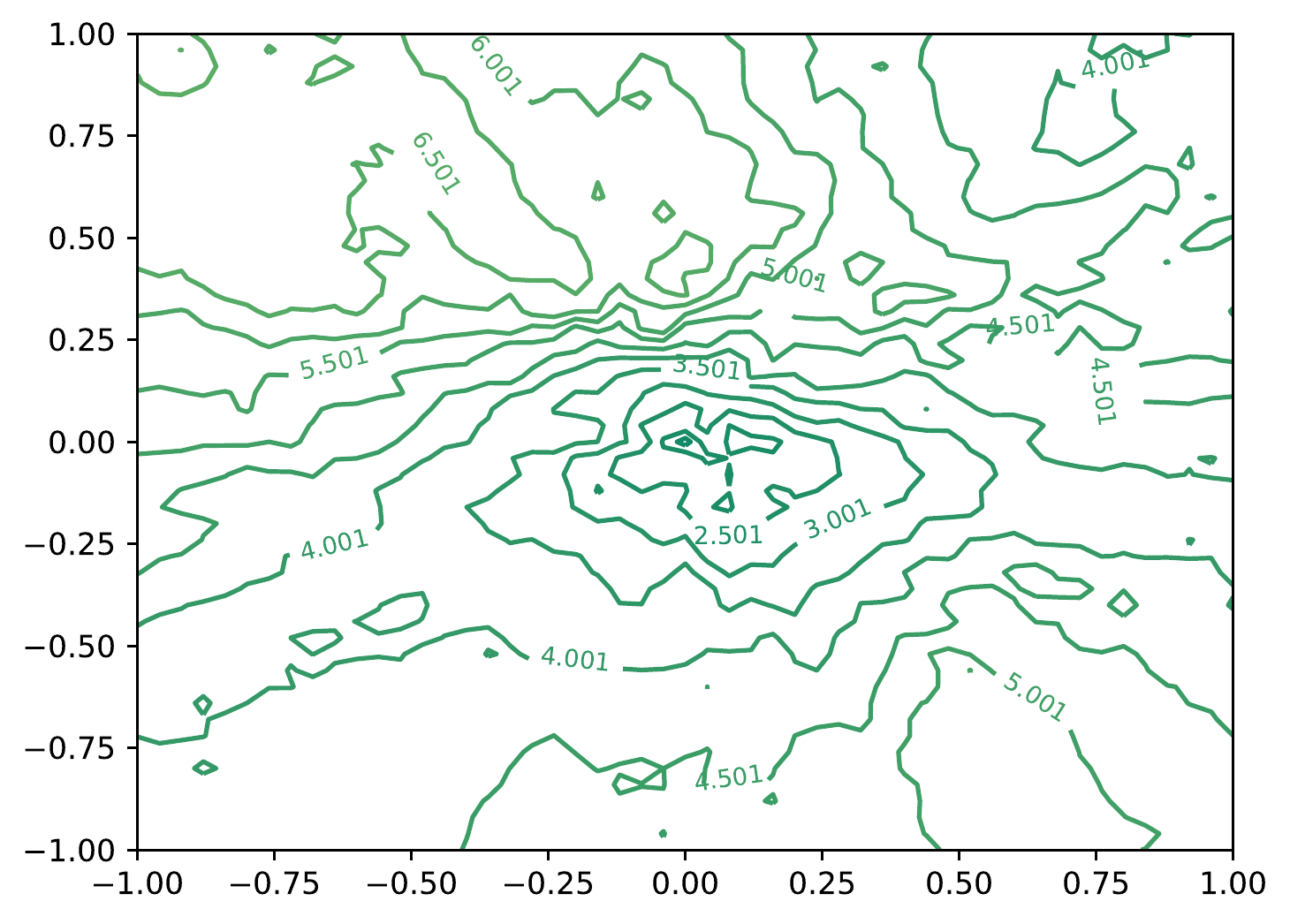}
    \centering
    \caption{Real2BinaryNet-18}
    \label{fig:loss landscape real2binarynet-18}
\end{subfigure}
\hfill
\begin{subfigure}[t]{.13\textwidth}
    \centering
    \includegraphics[width=22mm, height=22mm]{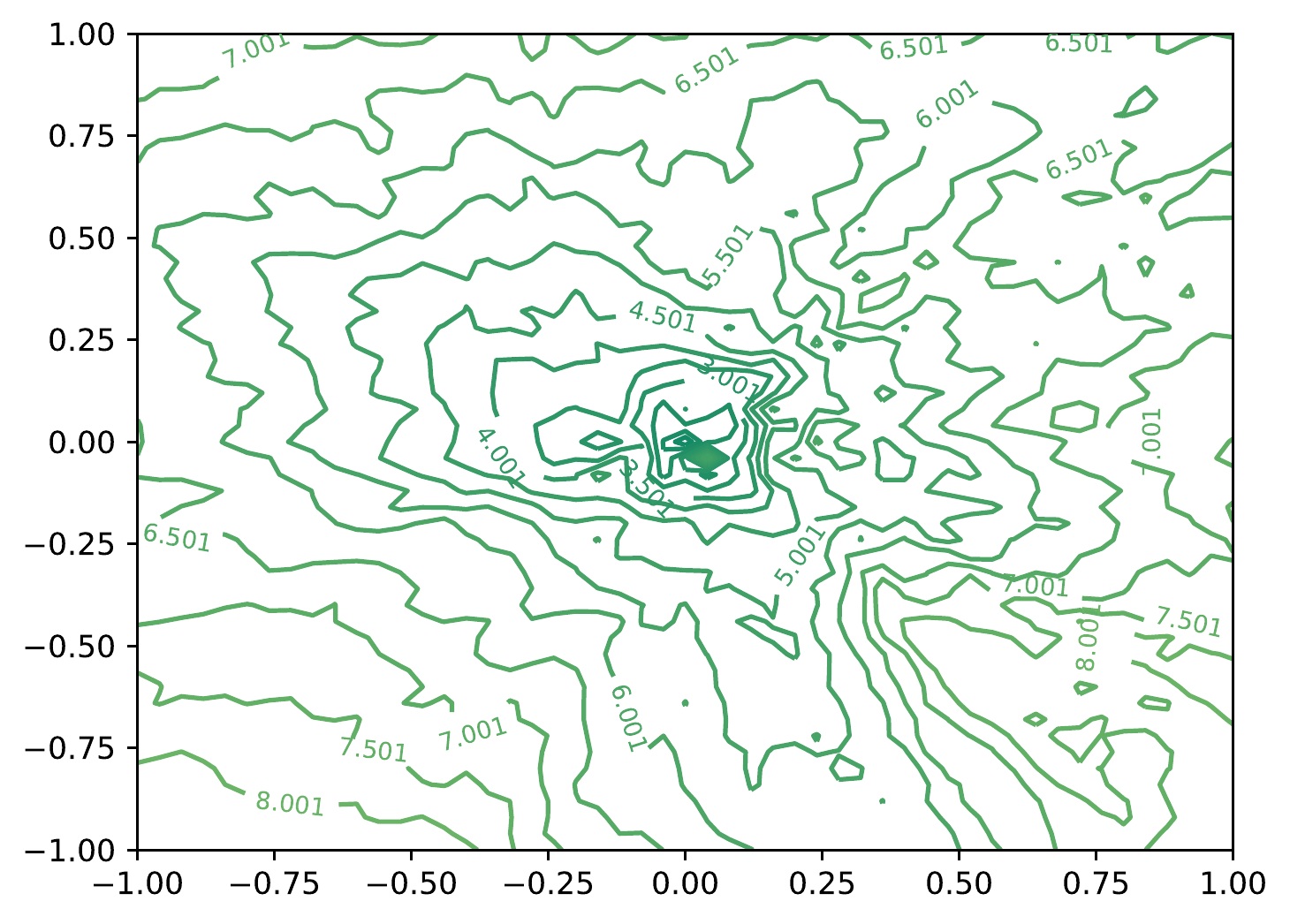}
    \centering
    \caption{BiRealNet-18 + Info-RCP (ours)}
    \label{fig:loss landscape birealnet-18 + info-rcp}
\end{subfigure}
\hfill
\begin{subfigure}[t]{.13\textwidth}
    \centering
    \includegraphics[width=22mm, height=22mm]{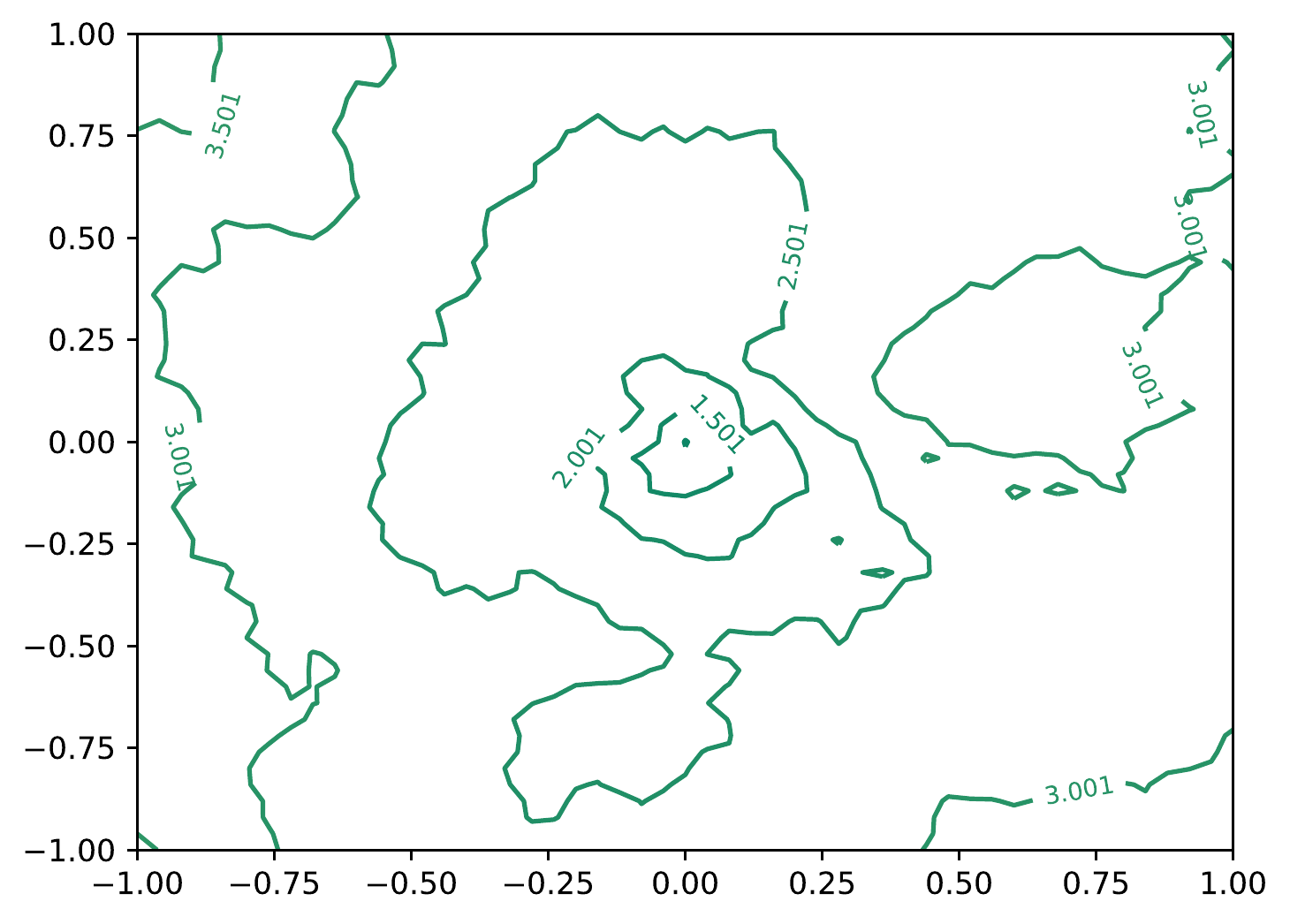}
    \centering
    \caption{BiRealNet-18 + ELM-Attention (ours)}
    \label{fig:loss landscape birealnet-18 + ELM Attention}
\end{subfigure}
\hfill
\begin{subfigure}[t]{.13\textwidth}
    \centering
    \includegraphics[width=22mm, height=22mm]{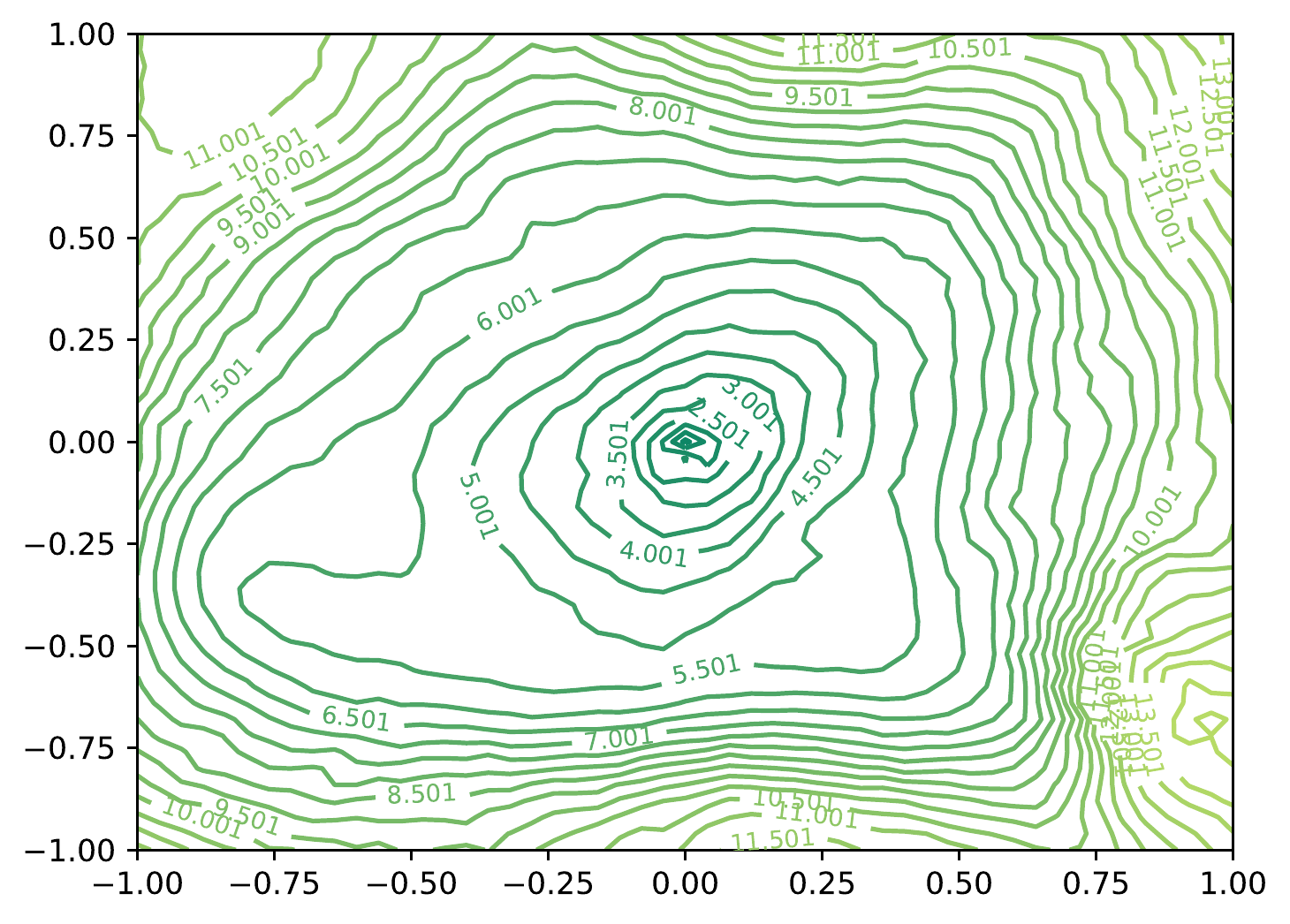}
    \centering
    \caption{BNext (ours)}
    \label{fig:loss landscape bnext}
\end{subfigure}
\hfill
\end{center}
\caption{The loss landscape of popular BNNs (contour line view) on CIFAR10 dataset. we follow the setting of \cite{li2018visualizing}.  ``Info-RCP'' and ``ELM-Attention'' are the module designs in this paper. The corresponding 3D views are available in the supplementary material.}
\label{fig:loss_landscape_visualization}
\end{figure*}

\begin{figure}[]
\captionsetup[subfigure]{justification=centering, font=tiny}
    {\begin{subfigure}[c]{.28\columnwidth}
        \includegraphics[height=26mm,clip=false]{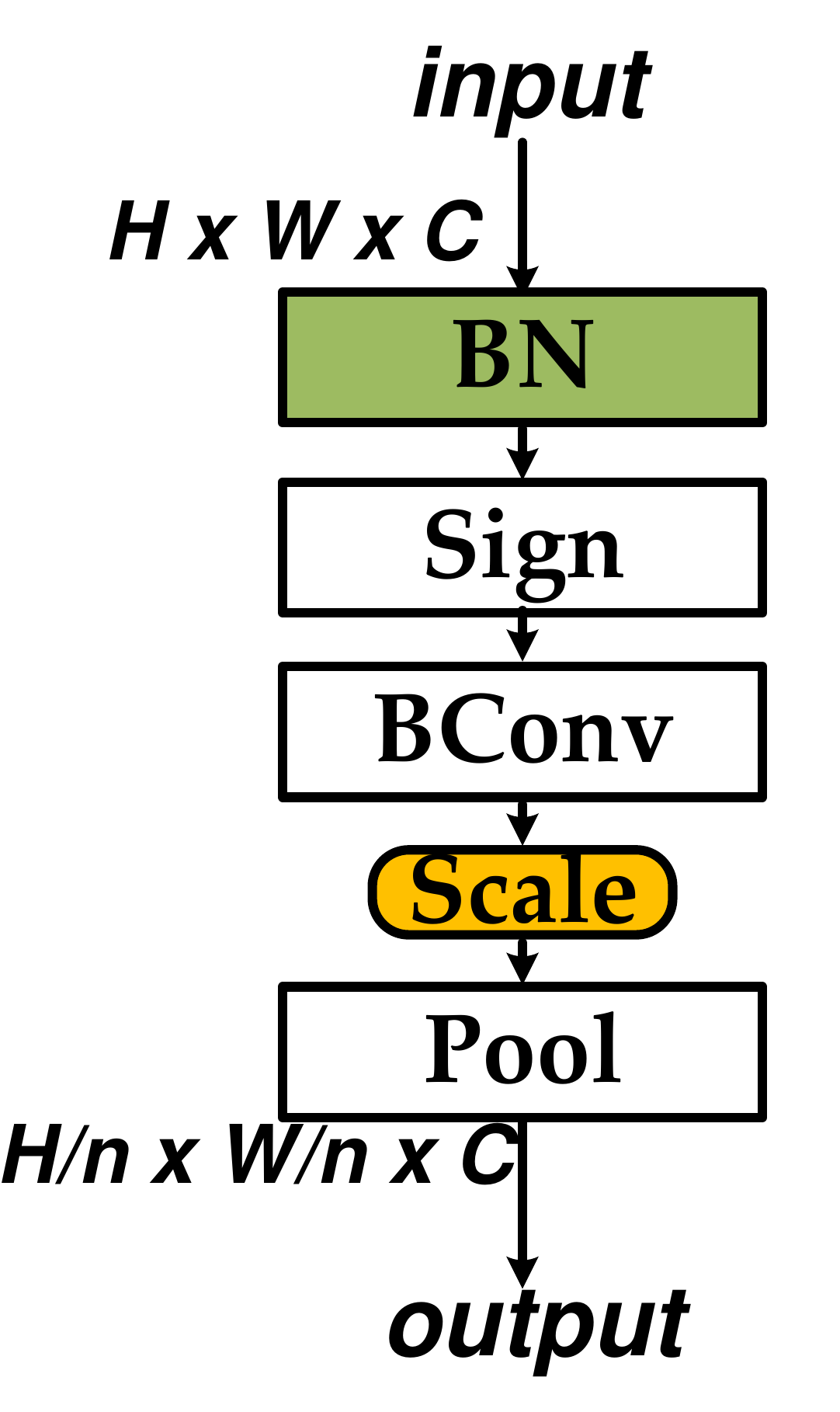}
        \caption{XNOR Net}
        \label{fig:xnor conv block}
    \end{subfigure}}
    {\begin{subfigure}[c]{.32\columnwidth}
        \includegraphics[ height=26mm,clip=false]{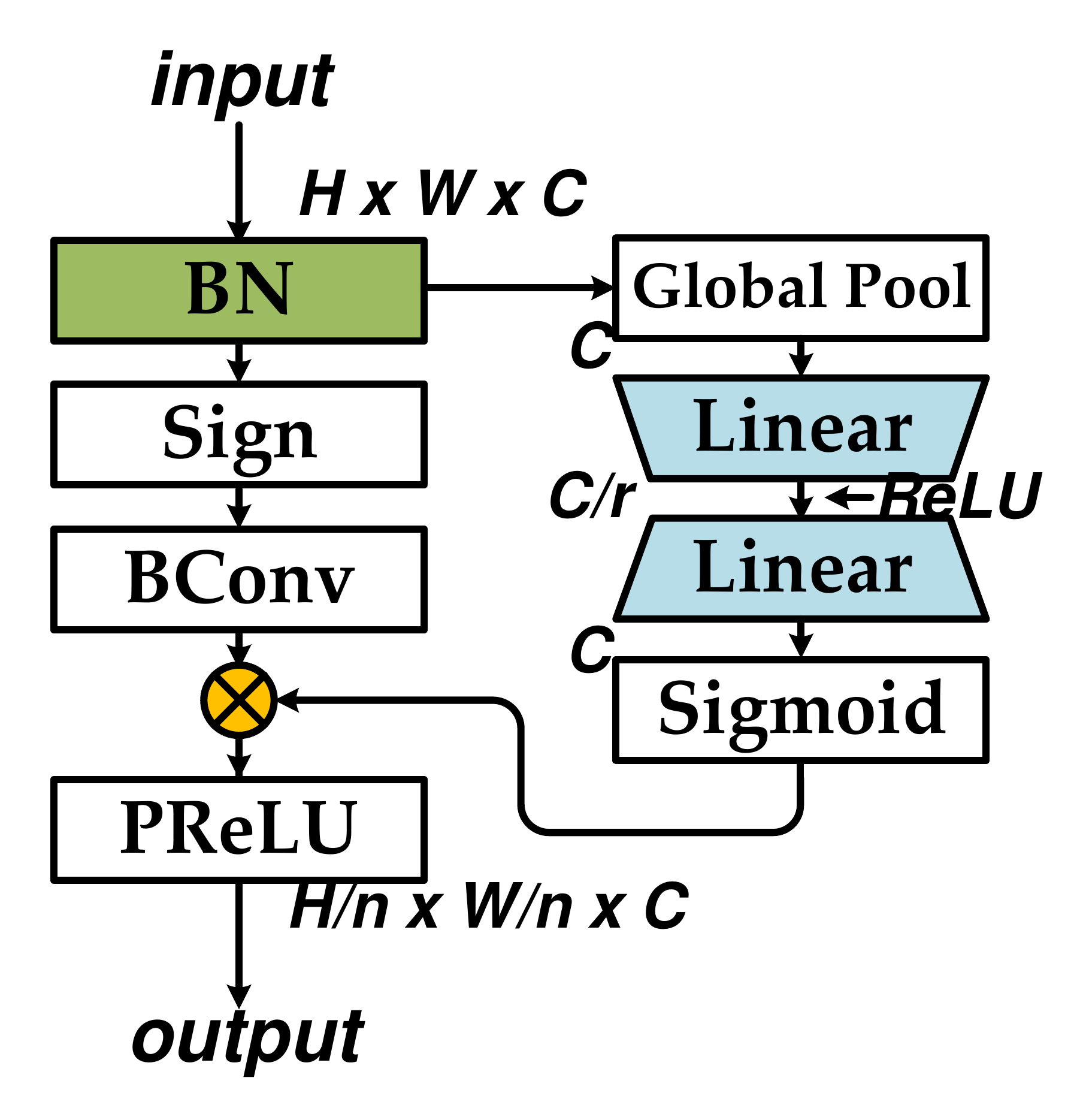}
        \caption{Real2BinaryNet}
        \label{fig:real2binary conv block}
    \end{subfigure}}
    {\begin{subfigure}[c]{.38\columnwidth}
        \includegraphics[height=26mm,clip=false]{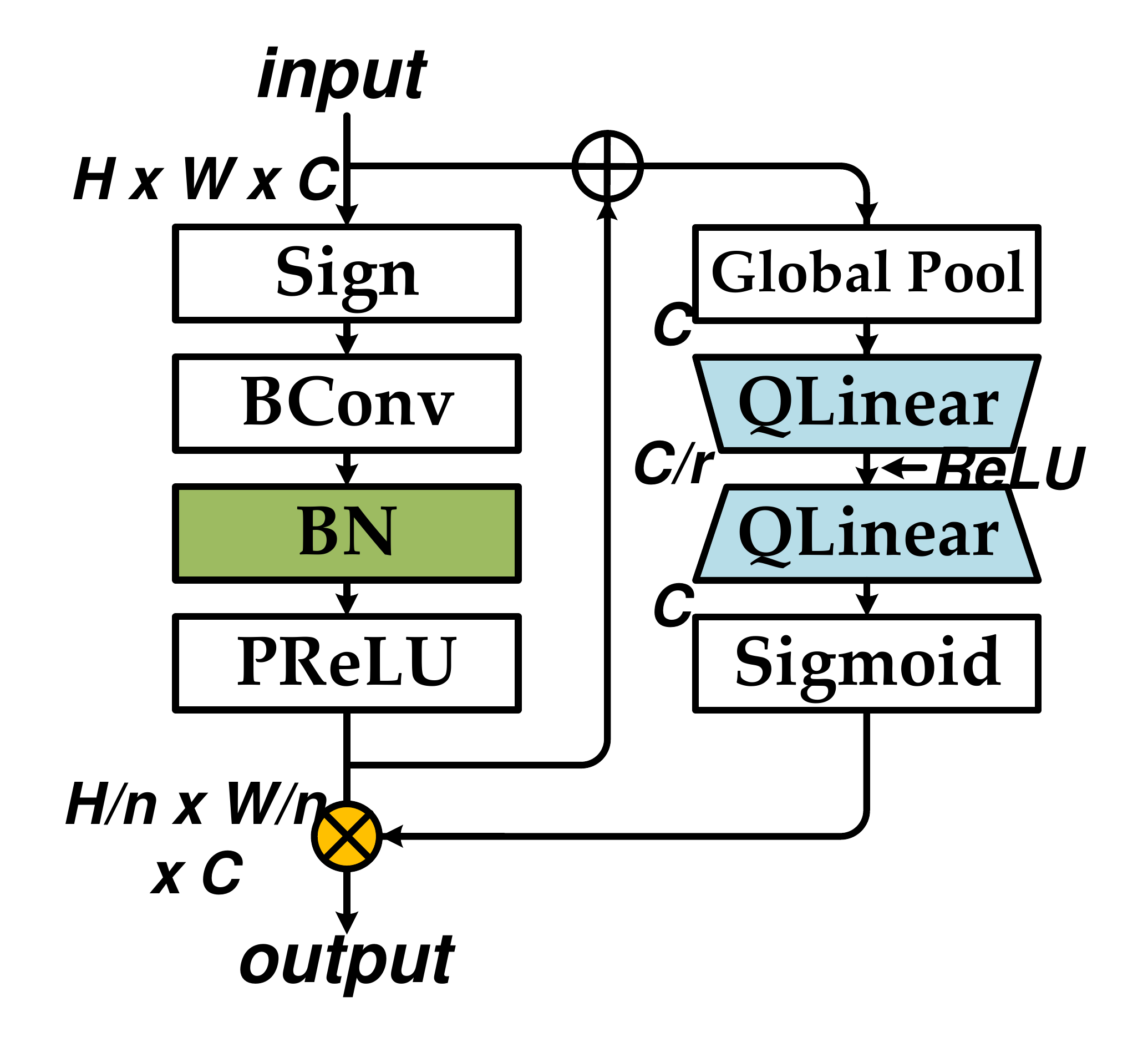}
        \caption{Info-RCP (ours)}
        \label{fig:infor-rcp}
    \end{subfigure}}
\caption{Binary convolution module design comparison among XNOR-Net \cite{rastegari2016xnor}, Real2BinaryNet \cite{real2binICLR20} and Ours.}
\label{fig:Conv Block Compare}
\end{figure}
\section{BNext Architecture Design}
\label{sec:macro architecture design}
In this section, we introduce the motivation and details of BNext architecture design. 
In section \ref{subsection:Visualizing Optimization Bottleneck of BNNs}, we visualize the loss landscape of the most popular binary neural networks and provide our observations and insights.
We then present the two core modules of BNext architecture design and the structure of the BNext model family in section \ref{Section:Building Optimization-Friendly BNNs}.



\subsection{Visualizing the Optimization Bottleneck}
\label{subsection:Visualizing Optimization Bottleneck of BNNs}
Binary neural networks are lightweight neural networks constructed from stacked 1-bit convolutions. 
In forward propagation, both the inputs $x_r^t$ and the proxy weights $w_r^t$ in each 1-bit convolution are binarized into ${\pm{1}}$ using the sign function before the dot product computation. 
In backward propagation, the sign function is approximated using the gradient of hardtanh function \cite{hubara2016binarized}, so that we can compute the gradient of $w_b^t$ and optimize the weights $w_r^t$. 
The gradient of $w_b^t$ is accumulated in the proxy weights $w_r^t$ during each iteration  \cite{courbariaux2016binarized}. This process can be mathematically formulated as follows:
\begin{equation}
\label{func. binarization}
\small
x_b^t, w_b^t = \mathrm{binarize}\left( x_r^t, w_r^t \right) =\begin{cases}
	+1&		if\ x_r^t, w_r^t\ge 0,\\
	-1&		\mathrm{otherwise.}\\
\end{cases}
\end{equation}
\begin{equation}
w_{r}^{t+1}=w_{r}^{t}-\alpha \cdot \frac{\partial \ell ^t}{\partial w_{r}^{t}},\ \ \ \frac{\partial \ell ^t}{\partial w_{r}^{t}}=\frac{\partial \ell ^t}{\partial w_{b}^{t}}\cdot \frac{\partial w_{b}^{t}}{\partial w_{r}^{t}},
\end{equation}
where ${\frac{\partial w_{b}^{t}}{\partial w_{r}^{t}}=1_{||w_{r}^{t}||\le c},}$ represents the Straight-Through-Estimation (STE) strategy. 
$r$, $b$, and $t$ denote real-valued variable, binary variable, and iteration number, respectively. 
Despite STE partially solving the non-differentiable problem, the sparse binary representation and gradient mismatch problem make BNNs suffer from serious accuracy degradation \cite{courbariaux2016binarized,rastegari2016xnor}. 
Various subsequent designs \cite{rastegari2016xnor,liu2018birealnet,real2binICLR20,bethge2020meliusnet,liu2020reactnet} have tried to solve the two core bottlenecks of BNNs: capacity degradation and optimization difficulty.  
To understand the optimization bottleneck in BNNs, we first focus on the most popular BNN architectures. 
Specifically, previous researchers have proposed several performance-beneficial structures in BNNs, such as double residual connection \cite{liu2018birealnet}, attention as introduced in Real2BinaryNet \cite{real2binICLR20, zhang2021pokebnn} and adaptive distribution reshaping \cite{liu2020reactnet}.
However, the shared insight behind these techniques is still unclear.

Liu \etal \cite{liu2021adam} studied the impact of training strategies and optimizers on BNNs by visualizing the loss landscape.
Inspired by this idea, we further use it as an indicator to analyze the most popular binary architectures, including BinaryNet \cite{courbariaux2016binarized}, BiRealNet \cite{liu2018birealnet}, Real2BinaryNet \cite{real2binICLR20}, and their full-precision counterpart ResNet-18 \cite{he2016deep}. 
We plotted the corresponding 2D loss landscapes (contour line view) for each network, as shown in Fig.~\ref{fig:loss_landscape_visualization}. 
By comparing Fig.~\ref{fig:loss landscape resnet18} and Fig.~\ref{fig:loss landscape binarynet-18}, we can see that directly binarizing the network \cite{courbariaux2016binarized} makes the loss landscape surface extremely discontinuous and rugged, the contour lines are unstructured and are unaligned in Fig.~\ref{fig:loss landscape binarynet-18}. 
The highly rugged landscape surface makes BNNs easier to converge into sub-optimal minima and more sensitive to inappropriate optimization processes. 
In Fig.~\ref{fig:loss landscape birealnet-18}, the double skip-connection design proposed in BiRealNet \cite{liu2018birealnet} alleviates the information sparsity problem to some extent and allows a more continuous information flow through bypass connections. 
Therefore, we can observe more structured and complete contours, although they still appear rugged.
Real2BinaryNet \cite{real2binICLR20} suggests reshaping the output of binary convolution in a data-driven manner by using a trainable attention module, which reduces the information bottleneck.
Compared to Fig.~\ref{fig:loss landscape binarynet-18} and \ref{fig:loss landscape birealnet-18}, Real2BinaryNet further flattens the area around the global minimum and the gap between each contour line in Fig.~\ref{fig:loss landscape real2binarynet-18} is more uniform, which means more robust to the initialization and optimization. 

Inspired by these findings in the visualization, we formulate the following three hypotheses:
(1) The activation binarization restricts the feature patterns available for forward propagation before each 1-bit convolution layer. 
(2) Properly reshaping the distribution after a binary convolution is essential for the feature modeling between adjacent 1-bit processing units. 
(3) An effective shortcut design enhances the information density in forward and backward propagation of BNNs. 
Although a lot of previous works are based on similar optimization ideas \cite{rastegari2016xnor, liu2018birealnet, real2binICLR20, liu2020reactnet, zhang2021pokebnn}, as seen from the loss landscape, their optimization effects can still be improved compared to the full-precision backbone Fig.~\ref{fig:loss landscape resnet18}. 
This conclusion motivates us to develop a better architectural design.

\begin{figure*}[]
\captionsetup[subfigure]{justification=centering}
\begin{center}
\includegraphics[width=\textwidth]{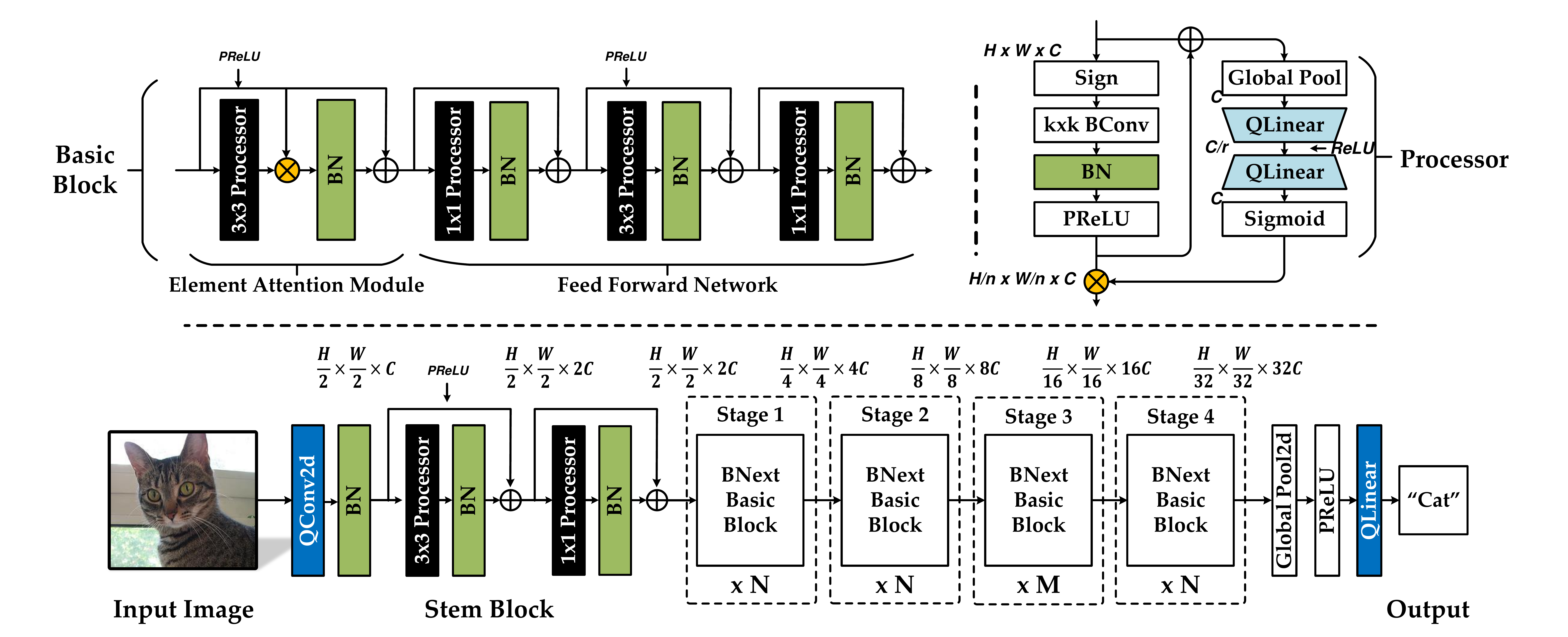}
\end{center}
\caption{The architecture of BNext. ``Processor'' is the basic binary convolution with improved Info-RCP design, ``BN'' represents batch normalization. ``C'' indicates the base width (channel number). ``N'' and ``M'' are the number of basic blocks in each stage respectively.}
\label{fig:architecture}
\end{figure*}

\subsection{Building Optimization-Friendly BNNs}
\label{Section:Building Optimization-Friendly BNNs}
This section introduces the core building components and their design ideas of our BNext network design.

\textbf{Binary Convolution with Information Recoupling} 
To facilitate the information propagation and alleviate possible bottlenecks between adjacent 1-bit processing units, we construct a new binary convolution module with adaptive information reshaping and coupling, as shown in Fig.~\ref{fig:infor-rcp}. 
During the forward pass, the output distribution of a binary convolution is adjusted with a subsequent Batch Norm and a PReLU layer. 
Next, we explicitly calibrate the information flow through the shift and scale design.
We first use a shortcut connection to fuse the input and output of the main branch [BConv-BN-PReLU] by element-wise additioning. 
The output is subsequently fed into a lightweight gap-aware Squeeze-and-Expand (SE) \cite{hu2018squeeze} branch, which is used to scale the main branch output. 
Specifically, both features before and after the main branch are considered as the input, which enables SE to perceive more information flow changes and to adaptively adjust the output distribution through learning.
This design is referred to as Information-Recoupling (Info-RCP) module in the rest of the paper.
Compared to XNOR-Net and Real2Binary-Net (Fig.~\ref{fig:xnor conv block} and \ref{fig:real2binary conv block}), our design offers a significantly more robust and flexible post-convolution reshaping mechanism.
Our ablation study (Table \ref{tab:ablation on module designs}) verifies the effectiveness of Info-RCP. 
Meanwhile, we find that the loss landscape also becomes more uniform and smooth (Fig.~\ref{fig:loss landscape birealnet-18 + info-rcp}), indicating that Info-RCP can effectively facilitate optimization.

\textbf{Basic Block Design with ELM-Attention}
Motivated by the analysis in Sec. \ref{subsection:Visualizing Optimization Bottleneck of BNNs} and the recent regularized architecture design ideas in the vision transformer \cite{vaswani2017attention,dosovitskiy2020image, han2021demystifying}, we investigate a novel basic block design with an enhanced bypass structure. 
As shown in Fig.~\ref{fig:architecture}, we stack multiple Info-RCP modules (Processor) to build the fundamental building block. 
Each Processor is followed by a BatchNorm layer. 
Then we use continuous residual connections to enclose each basic Processor [Info-RCP-BN], which relieves the information bottlenecks in forward propagation. 
Furthermore, we utilize an element-wise bypass multiplication to dynamically calibrate the output of the first 3$\times$3 Processor. 
The element-wise multiplication is helpful for the forward feature fusion and propagation in each block, which can substantially improve the loss landscape, as shown in Fig.~\ref{fig:loss landscape birealnet-18 + ELM Attention}.
Finally, by combining both Info-RCP and ELM-Attension we obtain a significantly better loss landscape, as shown in Fig.~\ref{fig:loss landscape bnext}.
Its smoothness is already very close to the full-precision backbone.


\textbf{BNext Family Construction} 
We build the BNext family by stacking basic blocks under different stage design strategies.
For a fair comparison with existing works, we apply the proposed block designs to the popular backbone MobileNetV1 and get the BNext variants by changing the block width and depth in each stage. 
The overall BNext family architectures are shown in Fig.~\ref{fig:architecture}.  
Specifically, we build four BNext variants with increasing capacity to test the scalability of BNext architecture.


Table \ref{tab:architecture information} shows the configuration of BNext networks.
$C$ indicates the width of input layer, $N$ and $M$ denote the stage depth ratios, as shown in Fig.~\ref{fig:architecture}. 
\begin{table}[]
\centering
\caption{BNext architecture overview. ``Q'' represents the quantization bit-width of weights and activations (W/A) in Fig. \ref{fig:architecture} for the first layer, the SE branch and the output layer, respectively.
}
\setlength\tabcolsep{4pt}
\footnotesize
\begin{tabular}{cccc}
\hline
Models  & Stage Ratios (N,M)  & Base Width (C) & Q (W/A) \\ \hline
BNext-T & 1:1:3:1       & 32         & 8/8-4/8-8/8     \\ \hline
BNext-S & 1:1:3:1       & 48         & 8/8-4/8-8/8     \\ \hline
BNext-M & 2:2:4:2       & 48         & 8/8-4/8-8/8     \\ \hline
BNext-L & 2:2:8:2       & 64         & 8/8-4/8-8/8     \\ \hline
\end{tabular}%
\label{tab:architecture information}
\end{table}

\section{Knowledge Distillation Strategy}

\subsection{Knowledge Complexity for Teacher Selection}
Knowledge Distillation (KD) is an essential optimization technique for BNNs \cite{real2binICLR20, liu2020reactnet,liu2021adam,zhang2021pokebnn}. 
To achieve the target of 80\% accuracy on ImageNet, we thus need to assess which teacher models provide the best supervision for BNNs. 
The only study that has studied the impact of teacher selection on high-accuracy BNN optimization is PokeBNN \cite{zhang2021pokebnn}, which uses a high precision VIT teacher instead of a ResNet-50 \cite{dosovitskiy2020image}. 
However, they report that the student model does not gain accuracy from a more accurate teacher, reflecting that just considering the teacher's accuracy is insufficient.
In our work, we did a more comprehensive investigation.
We observed different degrees of overfitting on BNext-T and BNext-L if we used stronger full-precision teachers and the standard KD.
Due to space limitations, we provide more details of the counter-intuitive overfitting in the Appendix (Section \ref{appendix: Counter Intuitive Overfitting}). 
To solve this issue, we propose a simple yet effective metric, \emph{Knowledge Complexity} (KC) incorporating accuracy and model compactness for teacher selection, and a diversified consecutive KD method.
\begin{equation}
    \mathcal{KC}(T) = \log{\left(1+\frac{\mathrm{Params}(T)}{\mathrm{Accuracy}(T) \cdot 10^6 + 10^{-8}}\right)},
\label{func. knowledge complexity}       
\end{equation}
where $T$ is the selected teacher model, $Params$ is the number of parameters and $Accuray$ the teacher's test accuracy. 
Generally speaking, KC defines the cost of teacher parameter-space complexity to the dark knowledge effectiveness. 
To the best of our knowledge, this is the first work to investigate this problem for BNNs and propose a practical solution.

The \emph{Knowledge Complexity} analysis of three popular DNN families and the corresponding teacher and student top-1 accuracy are shown in Fig.~\ref{fig:knowledge complexity}.
BNext-T has been used as the student in this study.
As common sense, we need a stronger teacher to pull the upper limit of students' accuracy.
However, as indicated by Fig.~\ref{fig:knowledge complexity}, a teacher with both high accuracy and high KC will readily cause the binary student to have the overfitting problem (significant drop in students' top-1 acc.).  
A possible reason is that training a binary student with a too-complex teacher at the initial training stage does not regard the tremendous discrepancy between the predicted distributions of the teacher and the student.
Similar observations have also been obtained in \cite{park2021prune}, where a pruned teacher works better than the original one.
Therefore, we suggest prioritizing teachers with lower KC for BNNs' KD training.

\subsection{Diversified Consecutive KD}
\label{subsec: knowledge distillation optimization}
We develop a diversified consecutive KD algorithm (Alg. \ref{Algorithm: Curriculum Knowledge Distillation}) to boost the performance of KD training further. 
This method consists of a gap-sensitive knowledge ensemble and a knowledge-boosting strategy.   

\begin{figure}[]
\captionsetup[subfigure]{justification=centering}
\begin{center}
\includegraphics[width=0.243\textwidth]{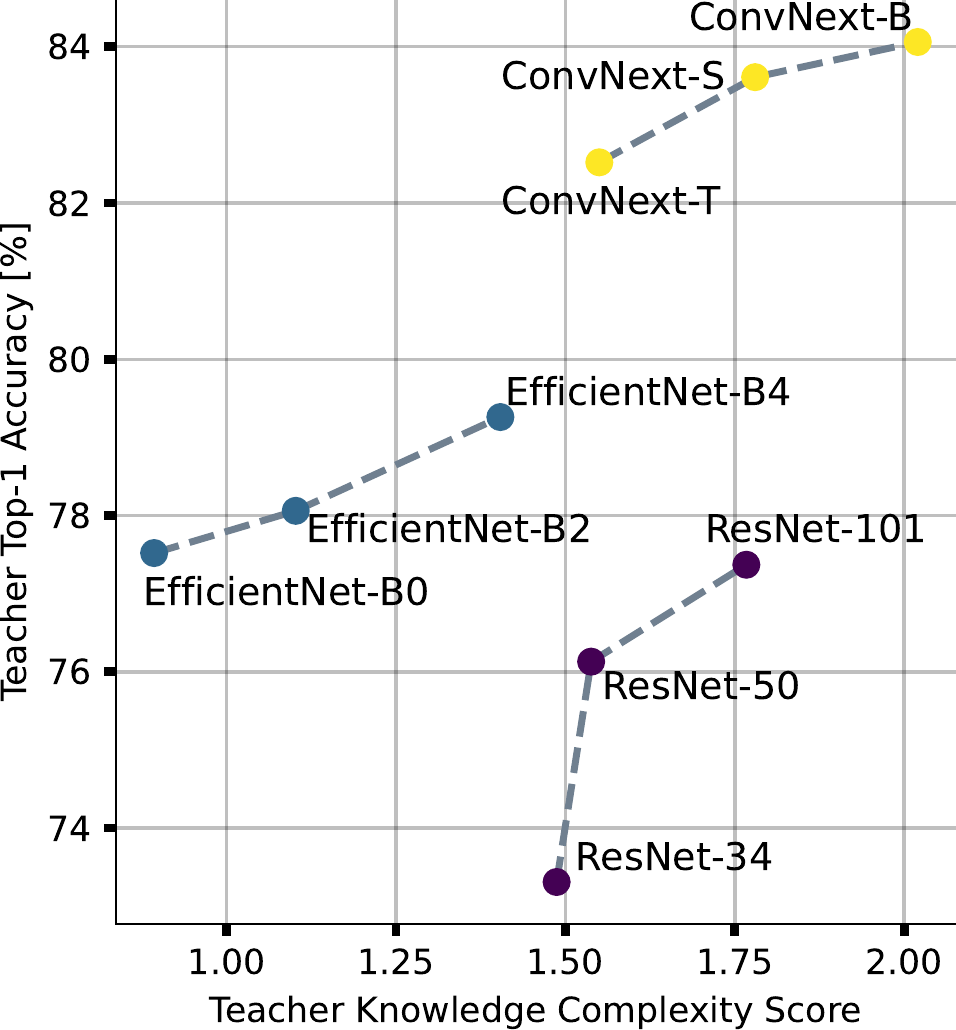}
\hfill
\includegraphics[width=0.227\textwidth]{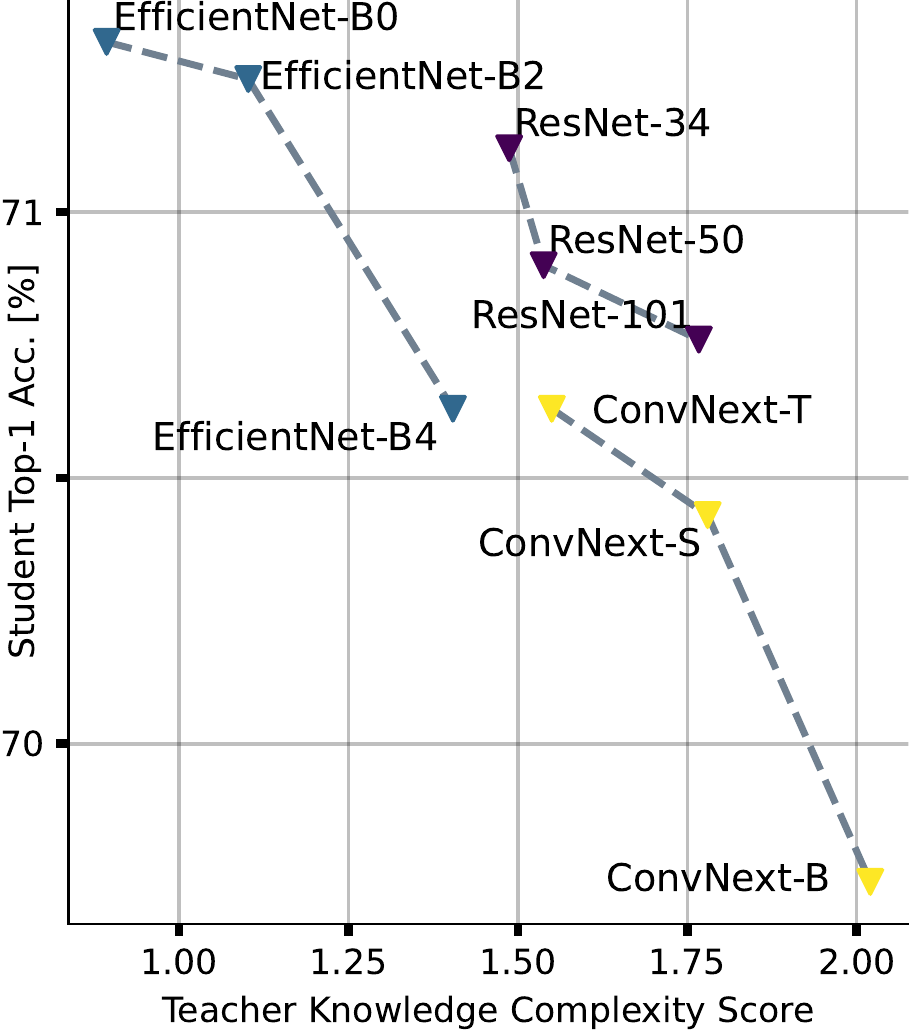}
\end{center}
\vskip -12pt 
\caption{Counterintuitive overfitting effects and the knowledge-complexity impacts under the selection of different teachers. The two y-axes indicate the teacher (left) and student (right) test Top-1 accuracy, evaluated on ImageNet with the input resolution 224×224. More details are available in supplementary materials.} 
\vskip -8pt 
\label{fig:knowledge complexity}
\end{figure}

\begin{algorithm}
\caption{Diversified Consecutive KD Algorithm}
\begin{algorithmic}
\Require A binary student $S_{\theta}$, a strong teacher $T_{\theta}^{s}$ and a group of assistant teachers \{$T_{\theta,1}^{w}, ..., T_{\theta,n}^{w}$\}. Training dataset $D$. Total epochs $N$. Initial assistant teacher index $m=1$. AMP quantization $Q$, provided by PyTorch. 

\For{${i}$ in epochs}: \Comment{start training}
    \For{$j$, $x$ in enumerate($D$)}: 
        \State 1.forward  $S_{\theta}(x), Q(T_{\theta}^{s})(x), Q(T_{\theta,m}^{w})(x)$,
        \State 2.calculate $\mathcal{L}(x, T_{\theta}^{s}, T_{\theta}^{w})$, Func. \ref{func. capacity aware KD}, 
        \State 3.backward propagation and updates, Func. \ref{func. binarization}.
    \EndFor
    \If{$i$\%$\left(\frac{N}{n}\right)$=0}: \Comment{switch assistant teacher $T_{\theta,m}^{w}$}
        \State m+=1,  \Comment{increase knowledge effectiveness}
    \EndIf
\EndFor             \Comment{end training}
\end{algorithmic}
\label{Algorithm: Curriculum Knowledge Distillation}
\end{algorithm}
\textbf{A Gap-Sensitive Knowledge Ensemble}
Our core design principle is to use the knowledge ensemble to increase the diversity of teacher knowledge. 
Moreover, the knowledge ensemble is not static but dynamically adapted regarding the difficulty of knowledge.
To guarantee a consistent accuracy upper bound, we use the same strong teacher ${T_{\theta}^s}$ during training.
We obtain additional guidance by adding another less confident assistant teacher ${T_{\theta}^w}$ to enhance the overall knowledge diversity.
The binary student ${S_{\theta}}$ consistently learns from two teachers.
The loss functions are mathematically formulated as follows:
\begin{multline}
    \mathcal{L}(x)=\sigma(x){KL}_{S_{\theta}, T_{\theta}^s}(x) + (1-\sigma(x)){KL}_{S_{\theta}, T_{\theta}^w}(x), \\
    \sigma(x) = \frac{e^{\mathcal{L}_{CE}(S_\theta, T_\theta^s)}}{\sum_i e^{\mathcal{L}_{CE}({S_\theta, T_\theta^i))}}}, \ \ \ \mathcal{L}_{CE}^i = -\sum_{x}S_\theta(x)\log(T_{\theta}^i(x)), 
\label{func. capacity aware KD}    
\end{multline}
where $CE$ and $KL$ indicate the Cross-Entropy and KL-Divergence loss \cite{shen2020meal}, respectively. 
We calculate the distance between student prediction and each teacher prediction for every forward pass. 
We then weigh each knowledge entry $KL_{S_{\theta}, T_{\theta}^i}$ using a Softmax term $\sigma(x)$. 
Consequently, our dynamic weighing and fusion method constrains the capacity-limited binary student ${S_{\theta}}$ to pay more attention to the more complex knowledge. 

\textbf{Knowledge-Boosting Strategy} 
\label{subsec: Knowledge-Boosting Strategy}
To further decrease the knowledge discrepancy in BNext-M/L optimization, we develop a knowledge-boosting strategy.
As aforementioned, the strong teacher ${T_{\theta}^s}$ is consistent throughout the training, but the assistant teacher is selected from a pre-defined candidate group.
The assistant teachers will join the KD process one by one in increasing order of accuracy and knowledge-complexity and have the same length of active time.
In this way, we use adapted assistants at different stages of training to adjust teacher knowledge, gaining better diversity. 
Binary students can thus receive more diverse supervision information, which have more robust regularization and can effectively solve the strong teacher overfitting problem shown in Fig.~\ref{fig:knowledge complexity}.
This is an important reason why BNext-M/L can achieve such high accuracy.
The teacher selection rule is based on knowledge complexity statistics. 
A detailed description of the assistant teacher group settings can be found in the supplementary material.
\begin{table*}[]
\centering
\caption{Performance comparison with modern DNN designs and SOTA binary designs on ImageNet.
``W/A'' indicates the bit-width of convolutions except for the input layer.
``QOPs'' is the INT-4/8 operations in total and ``$\dagger$" indicates results with post-training quantization.
In most cases, BNext performs better than models of similar (or larger) size - marked with (a) through (e).
}
\setlength\tabcolsep{5pt}
\footnotesize
\begin{tabular}{@{}rlcccc@{}p{24pt}@{}rlrccccc@{}}
\cline{2-6} \cline{9-15}
~ & \thead{Regular\\ Designs}     & \thead{W/A}   & \thead{OPs\\($10^{8}$)}   & \thead{\#Param\\(MB)}     & \thead{Top-1\\(\%)} &~&
~ & \thead{Binary\\Designs}       & \thead{BOPs\\$(10^{9})$}      & \thead{QOPs\\$(10^{6})$}       &\thead{FLOPs\\$(10^{8})$}      & \thead{OPs\\$(10^{8})$}       & \thead{\#Param\\(MB)} & \thead{Top-1\\(\%)} \\ \cline{2-6} \cline{9-15}
(a) & LQ-Net-18 \cite{zhang2018lq}       & 2/2   & -               & \phantom{00}8.4 & 64.9          &~&  ~  & BNN \cite{hubara2016binarized}           & 1.70      & -                & 1.20        & 1.47      & \phantom{0}4.2       & 42.2          \\
 ~  & LQ-Net-18 \cite{zhang2018lq}       & 4/4   & -               & \phantom{0}16.8 & 69.3          &~&  ~  & XNOR-Net \cite{rastegari2016xnor}        & 1.70      & -                & 1.20        & 1.47      & \phantom{0}4.2       & 51.2          \\
(b) & MobileNetV2 \cite{bai2019}         & 8/8   & \phantom{0}0.8  & \phantom{00}3.5 & 68.3          &~&  ~  & XNOR-Net++ \cite{bulat2019xnor}          & 1.70      & -                & 1.20        & 1.47      & \phantom{0}4.2       & 57.1          \\
(d) & ResNet-50 \cite{bai2019}           & 8/8   & \phantom{0}4.8  & \phantom{0}24.6 & 74.7          &~& (a) & Bi-RealNet-18 \cite{liu2018birealnet}    & 1.68      & -                & 1.39        & 1.65      & \phantom{0}4.2       & 56.4          \\
 ~  & ResNet-18 \cite{he2016deep}        & 32/32 & \phantom{}18.0  & \phantom{0}44.6 & 69.6          &~& (b) & Bi-RealNet-34 \cite{liu2018birealnet}    & 3.53      & -                & 1.39        & 1.94      & \phantom{0}5.1       & 62.2          \\
(c) & MobileNetV3 \cite{howard2019searching} & 32/32 & \phantom{0}2.2 & \phantom{0}21.6 & 75.2          &~&  ~  & Bi-RealNet-152 \cite{liu2020bi}          & 10.7      & -                & 4.48        & 6.15      & -                    & 64.5          \\
~ & ResNet-50 \cite{he2016deep}        & 32/32 & \phantom{}38.0  & \phantom{0}97.5 & 76.0          &~& (b) & MeliusNet-29 \cite{bethge2020meliusnet}  & 5.47      & -                & 1.29        & 2.14      & \phantom{0}5.1       & 65.8          \\
(e) & DeiT-S \cite{touvron2021training}  & 32/32 & \phantom{}46.0  & \phantom{0}83.9 & 79.7          &~&  ~  & MeliusNet-42 \cite{bethge2020meliusnet}  & 9.69      & -                & 1.74        & 3.25      & \phantom{}10.1       & 69.2          \\
(e) & RegNetY-4G \cite{radosavovic2020designing} & 32/32 & \phantom{}40.0  & \phantom{0}80.1 & 80.0          &~&  ~  & MeliusNet-59 \cite{bethge2020meliusnet}  & 18.30     & -                & 2.45        & 5.30      & \phantom{}17.4       & 71.0          \\
~ & Swin-T \cite{liu2021swin}          & 32/32 & \phantom{}45.0  & \phantom{}106.0 & 81.3         &~& (a) & Real2Binary-Net \cite{real2binICLR20}    & 1.67      & -                & 1.56        & 1.82      & \phantom{0}5.1       & 65.4          \\
 ~  & ConvNext-T \cite{liu2022convnet}   & 32/32 & \phantom{}45.0  & \phantom{}114.3 & 82.5      &~& (a) & ReActNet-BiR18 \cite{ReActNetGithub}     & 1.68      & -                & 1.63        & 1.89      & \phantom{0}4.2       & 65.9          \\    \cline{2-6}
 ~  &                                    &       &                 &                 &               &~& (b) & ReActNet-A \cite{liu2018birealnet}       & 4.82      & -                & 0.12        & 0.87      & \phantom{0}7.4       & 69.4          \\
 ~  &                                    &       &                 &                 &               &~& (b) & ReActNet-Adam \cite{Liu2019radam}        & 4.82      & -                & 0.12        & 0.87      & \phantom{0}7.4       & 70.5          \\ \cline{2-6}
 ~  &                                    &       &                 &                 &               &~& (c) & PokeBNN 1.75x \cite{zhang2021pokebnn}    & 11.03     & \phantom{0}21.3  & -           & 1.74      & \phantom{}16.3       & 76.8          \\ 
 ~  & \mth{Our Design\\ w/o Post-Quant.} & \mth{W/A} & \mth{OPs\\($10^{8}$)} & \mth{\#Param\\(MB)} & \mth{Top-1\\(\%)} 
                                                                                                     &~& (d) & PokeBNN 2.0x \cite{zhang2021pokebnn}     & 14.14     & \phantom{0}25.5  & -           & 2.27      & \phantom{}20.7       & 77.2          \\ \cline{2-6} \cline{9-15} 
 ~  & BNext-18 (ours)                    & 1/1   & \phantom{0}1.64  & \phantom{00}5.4 & 68.4         &~& (a) & BNext-18 $\dagger$ (ours)                & 1.68      & \phantom{}135.2  & -      & 0.43      & \phantom{0}2.2       & \textbf{67.9} \\
 ~  & BNext-T (ours)                     & 1/1   & \phantom{0}0.88  & \phantom{0}13.3 & 72.4         &~& (b) & BNext-T $\dagger$ (ours)                 & 4.82      & \phantom{0}13.4  & -      & 0.77      & \phantom{0}5.3       & \textbf{72.0} \\
 ~  & BNext-S (ours)                     & 1/1   & \phantom{0}1.90  & \phantom{0}26.7 & 76.1            &~& (c) & BNext-S $\dagger$ (ours)                 & 10.84     & \phantom{0}21.1  & -      & 1.72      & \phantom{}11.1       & \textbf{75.8}             \\ 
 ~  & BNext-M (ours)                     & 1/1   & \phantom{0}3.38  & \phantom{0}46.5 & 78.3         &~& (d) & BNext-M $\dagger$ (ours)                 & 20.09     & \phantom{0}24.3  & -      & 3.17      & \phantom{}20.4       & \textbf{77.9} \\ 
 ~  & BNext-L (ours)                     & 1/1   & \phantom{0}8.54  & \phantom{}106.1 & 80.6         &~& (e) & BNext-L $\dagger$ (ours)                 & 52.15     & \phantom{0}39.4  & -      & 8.19      & \phantom{}47.6       & \textbf{80.4} \\ \cline{2-6} \cline{9-15}
\end{tabular}
\label{Table:ImageNet SOTA Comparison}
\end{table*}

\section{Experiments}
In this section, we evaluate our model and methods on the ILSVRC12 ImageNet \cite{deng2009imagenet} and the CIFAR dataset.

\subsection{Experimental Setups}
\label{subsec:Experimental Settings and Datasets}
We use the AdamW \cite{loshchilov2017decoupled} optimizer ($\beta_{1} {=} 0.99$, $\beta_{2} {=} 0.999$) with an initial learning rate of $10^{-3}$ and weight decays of $10^{-3}$ and $10^{-8}$ for non-binary and binary parameters, respectively. 
We use 5 epochs of warmup for the learning rate and then reduce it with a Cosine Scheduler \cite{paszke2019pytorch}. 
We set the backward gradient clipping range of the STE (Func.~\ref{func. binarization}) as [-1.5, 1.5]. 
We use hard binarization (Func.~\ref{func. binarization}) for activation and progressive weight binarization as proposed in \cite{guo2021boolnet}.
Two different backbones, ResNet \cite{he2016deep} and MobileNetV1 \cite{howard2017mobilenets} are used for technique evaluations.
The model is trained on 8 Nvidia DGX-A100 GPUs. 

\textbf{ImageNet:} We train for an input resolution of 224x224 with a batch size of 512 for 512 epochs.
RandAugment (7, 0.5) \cite{cubuk2020randaugment} is used for enhanced data augmentation.
We use the Diversified Consecutive KD process described in Section.~\ref{subsec: knowledge distillation optimization} for BNext optimization with
ConvNext-Tiny \cite{liu2022convnet} as the strong teacher and [EfficientNet-(B0-B2-B4)\cite{tan2019efficientnet}, ConvNext-Tiny] as the candidate assistant teacher group.

\textbf{CIFAR:} We train the model with a batch size of 128 for 256 epochs.
We use the standard data augmentations random crop, random horizontal flip and normalization \cite{paszke2019pytorch}, and use cross entropy loss \cite{paszke2019pytorch} for optimization.
Results on CIFAR are the average of five runs. 


\subsection{Performance Evaluation}
\textbf{ImageNet:} BNext achieves a new state-of-the-art top-1 accuracy for BNNs on ImageNet classification (see Table \ref{Table:ImageNet SOTA Comparison}).
While most existing BNNs \cite{courbariaux2016binarized,rastegari2016xnor,liu2018birealnet,bethge2020meliusnet,real2binICLR20,liu2020reactnet, tu2022adabin}
are still less accurate than a 32-bit ResNet-18 (69.74\% top-1 accuracy) \cite{he2016deep}, BNext-XL pushes the upper boundary of BNNs to a crucial accuracy level of 80.57\%. 
With this result it surpasses most existing works by 10\% accuracy and achieves a result close to SOTA 32-bit designs, such as ConvNext \cite{liu2022convnet}, Swin Transformer \cite{liu2021swin} and RegNetY-4G \cite{radosavovic2020designing}. 
Compared to the previous SOTA design PokeBNN, our BNext-M$\dagger$ achieves 0.7\% higher accuracy with similar model size. 
The BNext architecture also relies on lower optimization requirements than PokeBNN family. 
PokeBNN is trained with a huge batch size of 8192 for 720 epochs \cite{zhang2021pokebnn} on 64 TPU-v3 chips, while BNext only uses a batch size of 512 for 512 epochs. 
Compared to the popular ReActNet \cite{liu2020reactnet} design, BNext-Tiny$\dagger$ achieves 2.6\% higher accuracy with 10M less operations. 
To further verify the effectiveness of the BNext design, we combine the proposed techniques with a ResNet-18 backbone.
We use standard data augmentation and only ResNet-34 as a teacher for a fair comparison. 
The BNext-18 achieves 68.4\% top-1 accuracy on ImageNet dataset, only 1.2\% lower than the original ResNet-18 \cite{he2016deep}. 
Meanwhile, the performance is higher than all the existing designs such as BNN \cite{courbariaux2016binarized}, XNOR-Net \cite{rastegari2016xnor}, BiReal-Net \cite{liu2018birealnet}, Real2BinaryNet \cite{real2binICLR20}, and ReActNet-BIR18 \cite{liu2018birealnet}. For the sake of fair comparison, our calculation of OPs  excludes  negligible floating-point counts as \cite{liu2018birealnet,bethge2020meliusnet,liu2020reactnet,liu2021adam,zhang2021pokebnn}.           

\textbf{CIFAR:} We further explore the generalization ability of BNext (with a ResNet-18 backbone) on the smaller dataset CIFAR10 (see Table.~\ref{exp:comparison cifar10}).
Compared to the latest designs such as AdaBNN \cite{tu2022adabin}, ReCU \cite{xu2021recu} and RBNN \cite{lin2020rotated}, BNext achieves the best performance.
It surpasses one of the most recent works AdaBNN by 0.5\% and closes the gap to 32-bit ResNet to 0.8\%.
The complete evaluation on CIFAR100 is provided in the supplementary material.  

\begin{table}[]
\centering
\small
\caption{A comparison of state-of-the-art BNNs on CIFAR10.}
\resizebox{\columnwidth}{!}{%
\setlength\tabcolsep{3pt}
\begin{tabular}{c|c|c|c}
\hline
Method      & W/A (Bitwidth)   & Top-1 Acc {[}\%{]} & Top-5 Acc {[}\%{]}\\ \hline
Baseline (ResNet-18)    & 32/32 & 94.8  & -                \\
RAD  \cite{ding2019regularizing}       & 1/1   & 90.5    & -            \\
IR-Net \cite{qin2020forward}      & 1/1   & 91.5    & -            \\
RBNN  \cite{lin2020rotated}      & 1/1   & 92.2     & -           \\
ReCU  \cite{xu2021recu}       & 1/1   & 92.8    & -            \\
AdaBNN \cite{tu2022adabin}     & 1/1   & 93.1     & -           \\
BNext (ResNet-18) (ours) & 1/1   & \textbf{93.6 ($\pm0.12$)}    & 98.8 ($\pm0.03$)            \\ \hline
\end{tabular}
}
\label{exp:comparison cifar10}
\end{table}

\begin{table}[]
\centering
\setlength\tabcolsep{7pt}
\footnotesize
\caption{The ablation study of Info-RCP and ELM-Attention on ImageNet, showing the top-1/top-5 accuracy.}
\label{tab:ablation on module designs}
\begin{tabular}{c|cc}
\hline
Method                & w/o ELM-Attention  & w/ ELM-Attention   \\ \hline
w/o Info-Recoupling        & 62.43/83.75           & 64.61/85.49                \\ 
w/ Info-Recoupling    & 64.91/85.52           & 65.02/85.73                \\ \hline
\end{tabular}
\end{table}


\subsection{Ablation Study}
\label{Ablation Study}
We evaluate the impacts of the proposed techniques in this paper in a detailed ablation study on the proposed module designs,  optimization schemes, data augmentation strategies and post quantization impacts.
We use BNext-Tiny for all experiments with only standard data augmentation and cross-entropy loss for model training (unless stated otherwise).  

\textbf{Module Designs.}
We conduct an ablation study on the proposed Info-RCP and ELM-Attention modules in Table.~\ref{tab:ablation on module designs}.
The baseline model without both structures only achieves 62.43\% on ImageNet.
Adding only ELM-Attention to BNext-Tiny increases the accuracy to 64.91\%.
Adding only Info-RCP to BNext-Tiny increases the accuracy to 64.61\%.
Adding both designs, achieves the highest accuracy of 65.02\%.
Thus, both designs enhance the representational capacity, which confirms the expectations raised by the loss landscape comparison in Fig.~\ref{fig:loss_landscape_visualization}.


\textbf{Optimization Schemes.}
We evaluate the effectiveness of all optimization schemes used in our experiments in Table \ref{exp:ablation training schemes}.
The traditional knowledge distillation strategy wth ResNet-101 \cite{liu2020reactnet} as teacher is utilized as backbone. Replacing the teacher with $KC(x)$ (Func. \ref{func. knowledge complexity}) based teacher selection (Efficient-B0) pushes the accuracy by 0.57\%.
This reveals the importance of teacher selection in binary network optimization.
Using a gap-sensitive ensemble (Func. \ref{func. capacity aware KD}) of Efficient-B0 and Efficient-B2 gives an extra 0.05\% improvement. A further 0.08\% accuracy gain is achieved by adding Rand Augmentation \cite{cubuk2020randaugment}. Extending the training epochs to 512 pushes the accuracy to 72.40\%.
\begin{table}[]
\centering
\caption{
The ablation study on the effects of adding the different training strategies on ImageNet.}
\setlength\tabcolsep{4pt}
\footnotesize
\begin{tabular}{c|ccc}
\hline
Method                                                   & Epochs &  Top-1 {[}\%{]} & Top-5 {[}\%{]} \\ \hline
\multicolumn{1}{c|}{+ KD }        & 128      & 70.76        & 89.64        \\
\multicolumn{1}{c|}{+ KC-Based $T(x)$ Selection} & 128      & 71.33 (+0.57)        & 90.02 (+0.38)       \\
\multicolumn{1}{c|}{+ Gap-Sensitive KD} & 128  & 71.38 (+0.05) & 90.04 (+0.02) \\
\multicolumn{1}{c|}{+ Rand Augmentation \cite{cubuk2020randaugment}}   & 128  & 71.46 (+0.08)  & 90.06 (+0.02) \\ 
\multicolumn{1}{c|}{+ Long Training}   & 512  & 72.40 (+0.94)  & 90.64 (+0.58) \\ 
\hline
\end{tabular}%
\label{exp:ablation training schemes}
\end{table}

\begin{table}[]
\centering
\caption{The ablation study on popular data augmentation strategies on ImageNet. BNext-Tiny is selected as the baseline.}
\setlength\tabcolsep{2pt}
\footnotesize
\begin{tabular}{c|ccc}
\hline
\multicolumn{1}{c}{Data Augmentation}     & Epochs & Top-1 (\%) & Top-5 (\%) \\ \hline
\multicolumn{1}{c|}{Baseline}                      & 128 & 65.02 & 85.73 \\
\multicolumn{1}{c|}{Mixup \cite{zhang2017mixup}}              & 128 & 65.83 \phantom{0}(+0.81) & 86.16 \phantom{0}(+0.43) \\
\multicolumn{1}{c|}{Cutmix \cite{yun2019cutmix}}             & 128 & 64.51 \phantom{0}(-0.51)  & 85.32 \phantom{0}(-0.41)\\
\multicolumn{1}{c|}{Repeat Augment \cite{hoffer2020augment}}        & 128 & 53.74 (-11.28)& 75.65 (-10.08) \\
\multicolumn{1}{c|}{Rand Augment \cite{cubuk2020randaugment}} & 128 & \textbf{67.51 \phantom{0}(+2.49)} & \textbf{87.28 \phantom{0}(+1.55)} \\
\multicolumn{1}{c|}{Rand Augment \cite{cubuk2020randaugment}, Mixup \cite{zhang2017mixup}} & 121    & 66.51 \phantom{0}(+1.49) & 86.77 \phantom{0}(+1.04)\\ \hline
\end{tabular}%
\label{exp:ablation data augmentation}
\end{table}

\textbf{Data Augmentations.}
We verify the effectiveness of existing popular data augmentations such as Mixup \cite{zhang2017mixup}, Cutmix \cite{yun2019cutmix}, Augment-Repeat \cite{hoffer2020augment} and Rand Augmentation \cite{cubuk2020randaugment}. The results are shown in Table. \ref{exp:ablation data augmentation}.
Most of the popular choices  \cite{dosovitskiy2020image,liu2022convnet, liu2021swin} are harmful for the BNext model optimization.
For example, the Augment-Repeat \cite{hoffer2020augment} decreases the Top-1 accuracy by 10.08\%.
Both Rand Augmentation and Mixup improve the generalization separately, but combining them produces sub-optimal results.
Applying only Rand Augmentation achieves the best improvement of 2.49\%.
\begin{table}[]
\centering
\caption{Impact of post-training quantization on the first input layer, the last layer, and the SE branch in BNext Design. The best and second-best quantization results are highlighted.} 
\setlength\tabcolsep{3pt}
\footnotesize
\begin{tabular}{c|cccccc}
\hline
\multicolumn{1}{c|}{I-S-E-O (W/A)}           & BNext-18  & BNext-T & BNext-S & BNext-M & BNext-L  \\ \hline
\multicolumn{1}{c|}{32/32}             & 68.37          & 72.36       & 76.06     &  78.27        & 80.57    \\ \hline
\multicolumn{1}{c|}{8/8-8/8-8/8-8/8}   & \textbf{68.38}          & \textbf{72.36}       &   \textbf{76.05}      & \textbf{78.11}        & \textbf{80.47}    \\ 
\multicolumn{1}{c|}{8/8-4/8-4/8-8/8}   & \textbf{67.94}          &   \textbf{72.04}      &  \textbf{75.75}       & \textbf{77.97}        & \textbf{80.37}    \\
\multicolumn{1}{c|}{8/8-4/8-4/4-8/8}   & 67.91          & 71.95        &   75.61       & 77.93        & 80.26   \\
\multicolumn{1}{c|}{8/8-4/4-4/4-8/8}   & 67.29          & 71.72       &  75.50       & 77.79        & 79.52   \\
\hline
\end{tabular}
\label{table: ablation quantization}
\end{table}

\textbf{Quantization Impacts.} 
We apply post-training quantization to the input convolution layer, the output fully-connected layer and the layers in each SE branch, Table. \ref{table: ablation quantization}. 
Without quantization, BNext-L reaches the top-1 accuracy of 80.57\%.
Applying 8-bit quantization to weights and activations of all layers only marginally reduces accuracy to 80.47\%.
Further, decreasing the bit-width of weights in SE branch to 4 bits (with 8-bit activations) achieves the final BNext-L model whose classification performance 80.37\% Top-1 accuracy is still higher than the 80\% mark. 
\section{Conclusion}
In this paper, we proposed the first binary neural network architecture which achieves 80.57\% Top-1 accuracy on ImageNet, BNext.
This is achieved by enhancing the optimization process with the novel Info-RCP and ELM-Attention modules.
The smoother loss landscape and enhanced optimization allows better fitting to the train data and even shows slight overfitting, which was not possible with previous BNN architectures.
To increase the generalization and reduce any overfitting, we propose \emph{Diversified Consecutive KD} and discover several interesting phenomena in the process.
The highly accurate results on ImageNet demonstrate that BNext is sufficiently effective as a strong feature extractor for representation learning.

{\small
\bibliographystyle{ieee_fullname}
\bibliography{cvpr2023}
}

\clearpage
\begin{figure*}[]
\captionsetup[subfigure]{justification=centering, font=footnotesize}
\begin{center}
\begin{subfigure}[t]{.242\textwidth}
    \centering
    \includegraphics[width=42mm, height=42mm]{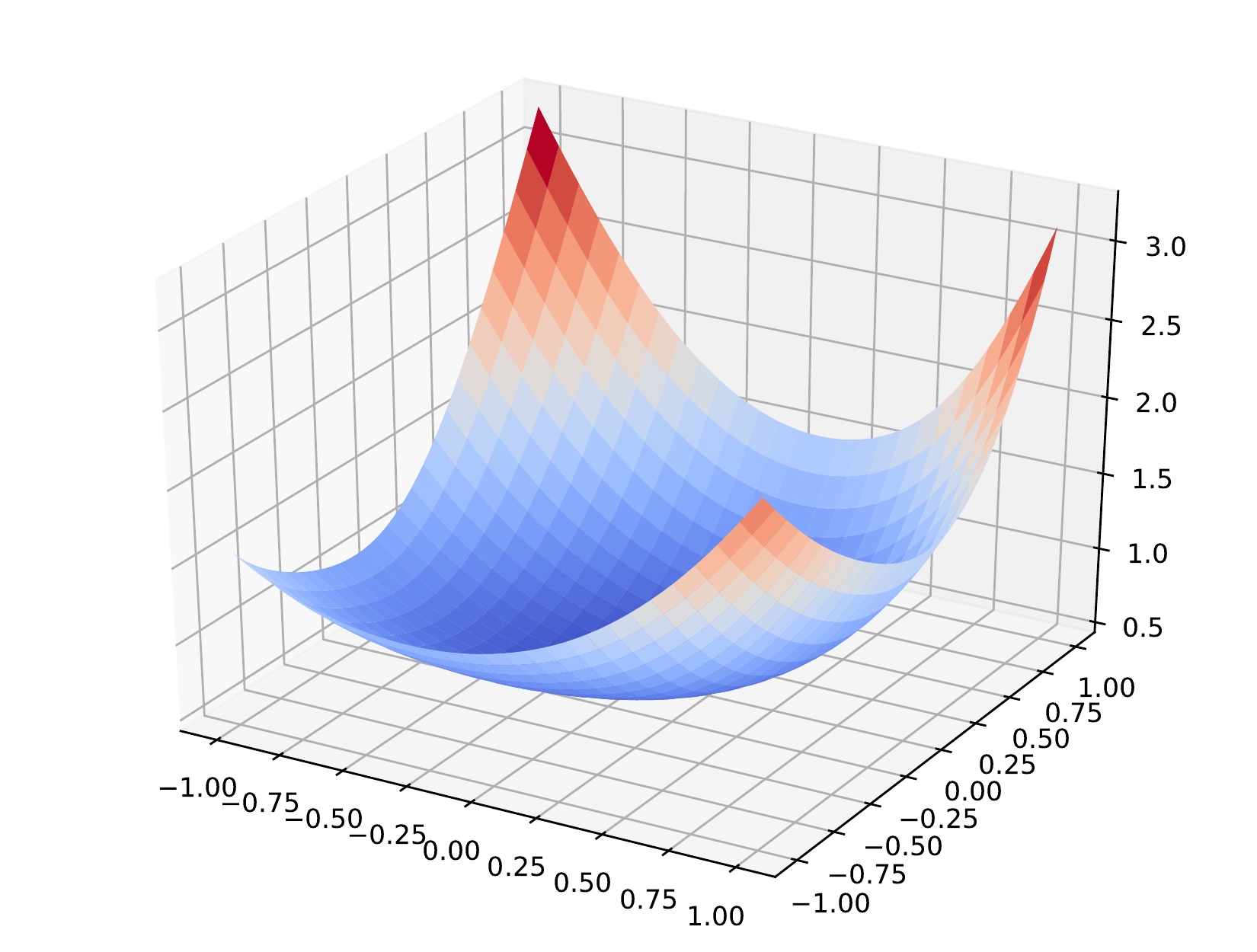}
    \centering
    \caption{ResNet-18}
    \label{fig:loss landscape resnet18 3d}
\end{subfigure}
\hfill
\begin{subfigure}[t]{.242\textwidth}
    \centering
    \includegraphics[width=42mm, height=42mm]{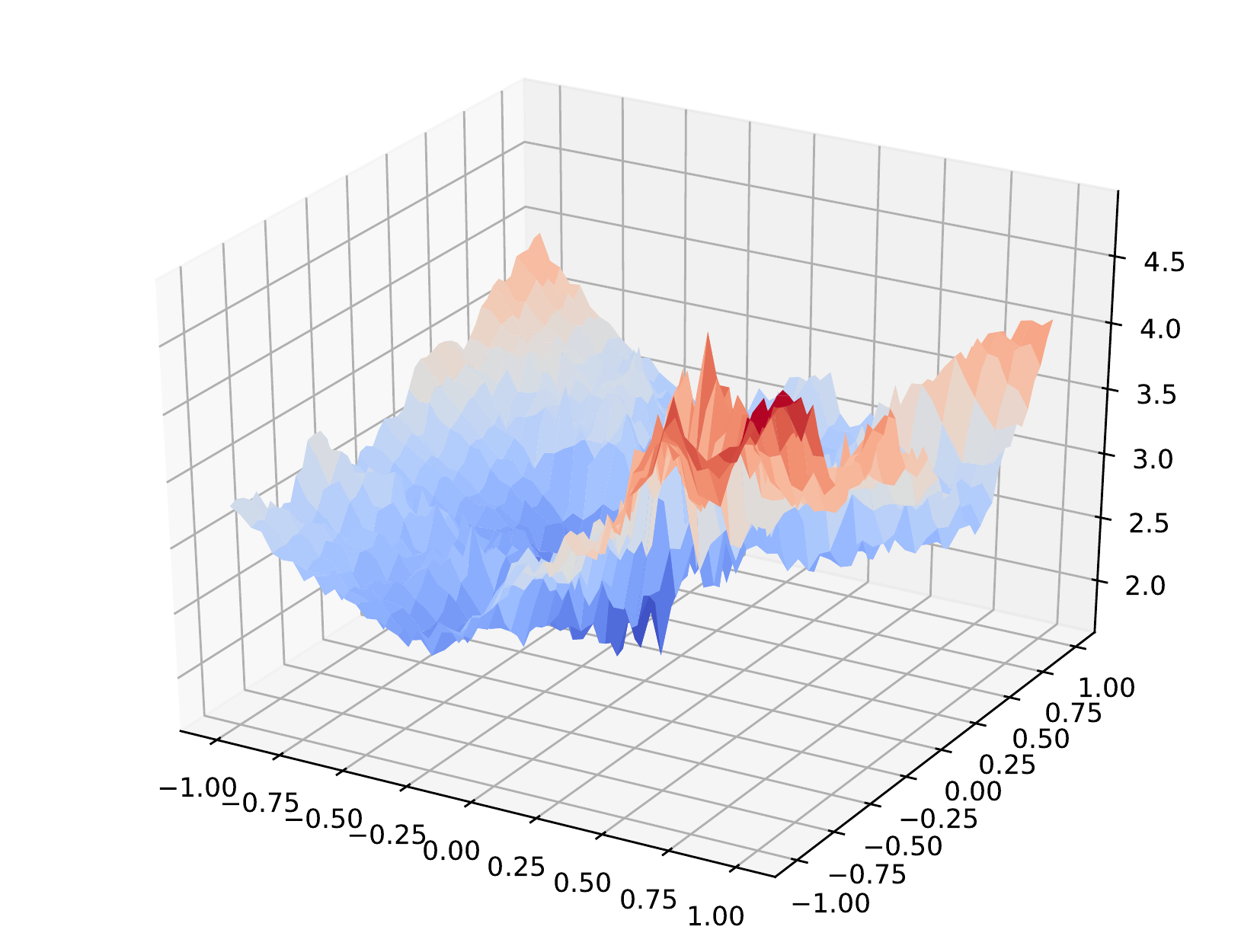}
    \centering
    \caption{BinaryNet-18}
    \label{fig:loss landscape binarynet-18 3d}
\end{subfigure}
\hfill
\begin{subfigure}[t]{.242\textwidth}
    \centering
    \includegraphics[width=42mm, height=42mm]{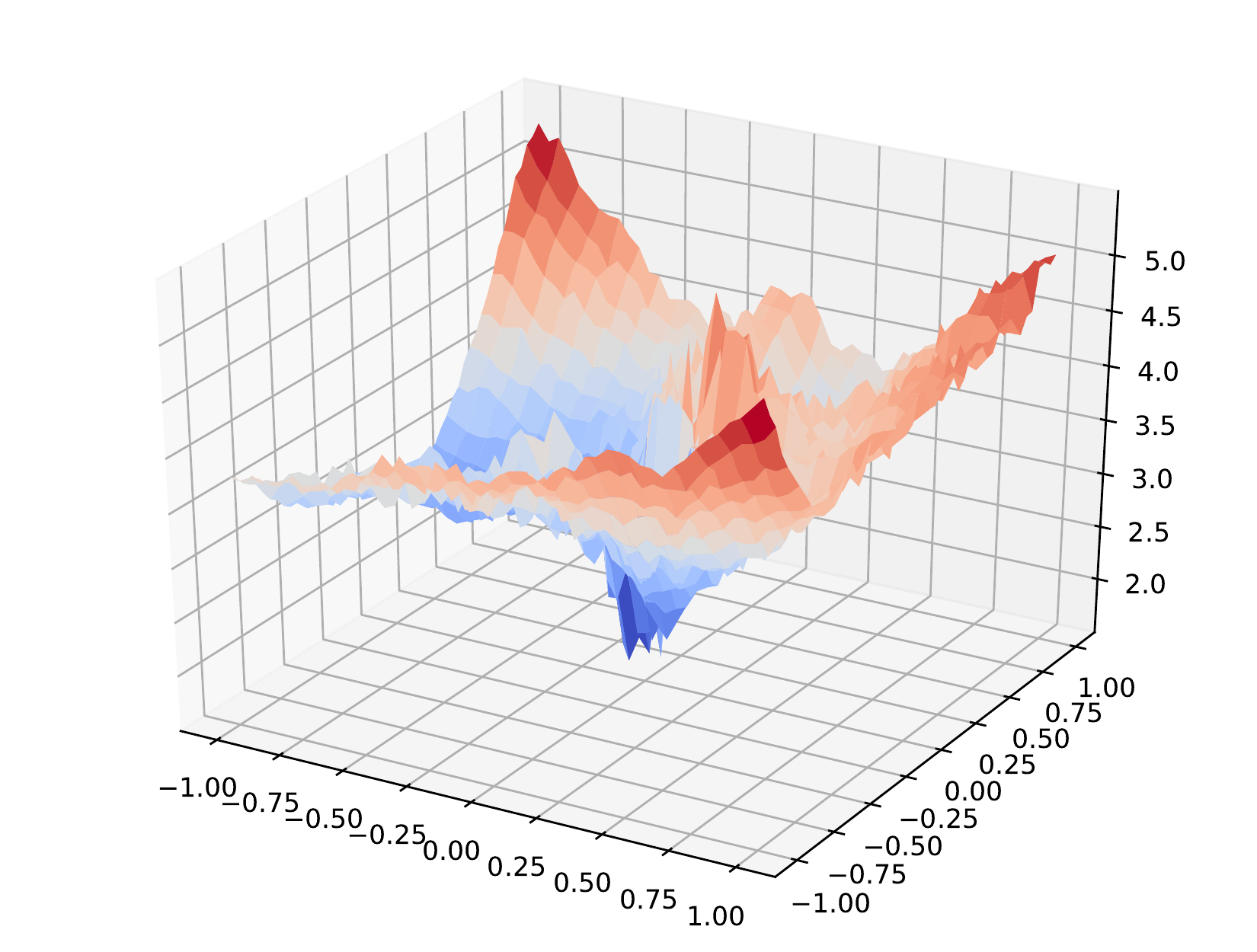}
    \centering
    \caption{BiRealNet-18}
    \label{fig:loss landscape birealnet-18 3d}
\end{subfigure}
\hfill
\begin{subfigure}[t]{.242\textwidth}
    \centering
    \includegraphics[width=42mm, height=42mm]{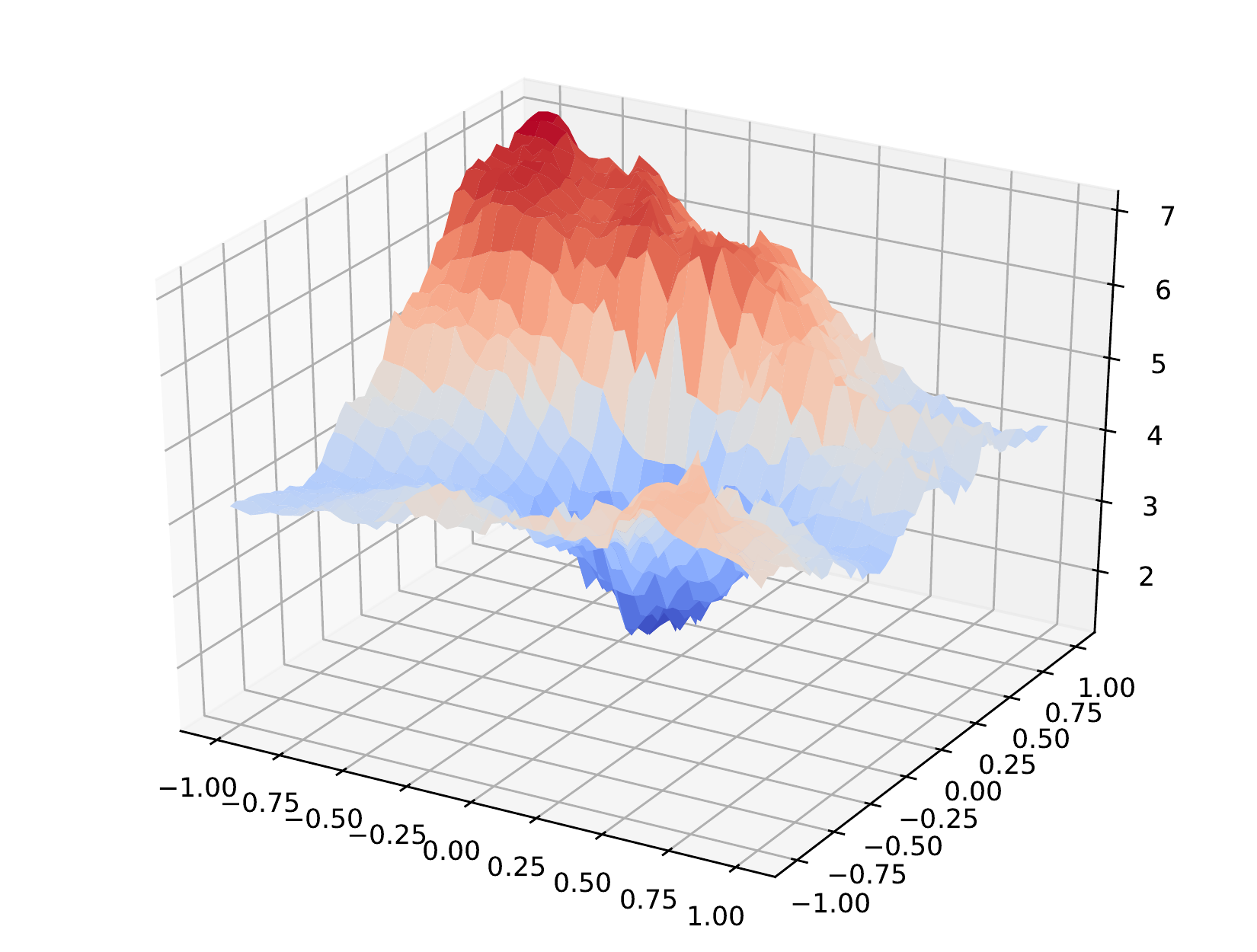}
    \centering
    \caption{Real2BinaryNet-18}
    \label{fig:loss landscape real2binarynet-18 3d}
\end{subfigure}
\hfill
\begin{subfigure}[t]{.3\textwidth}
    \centering
    \includegraphics[width=42mm, height=42mm]{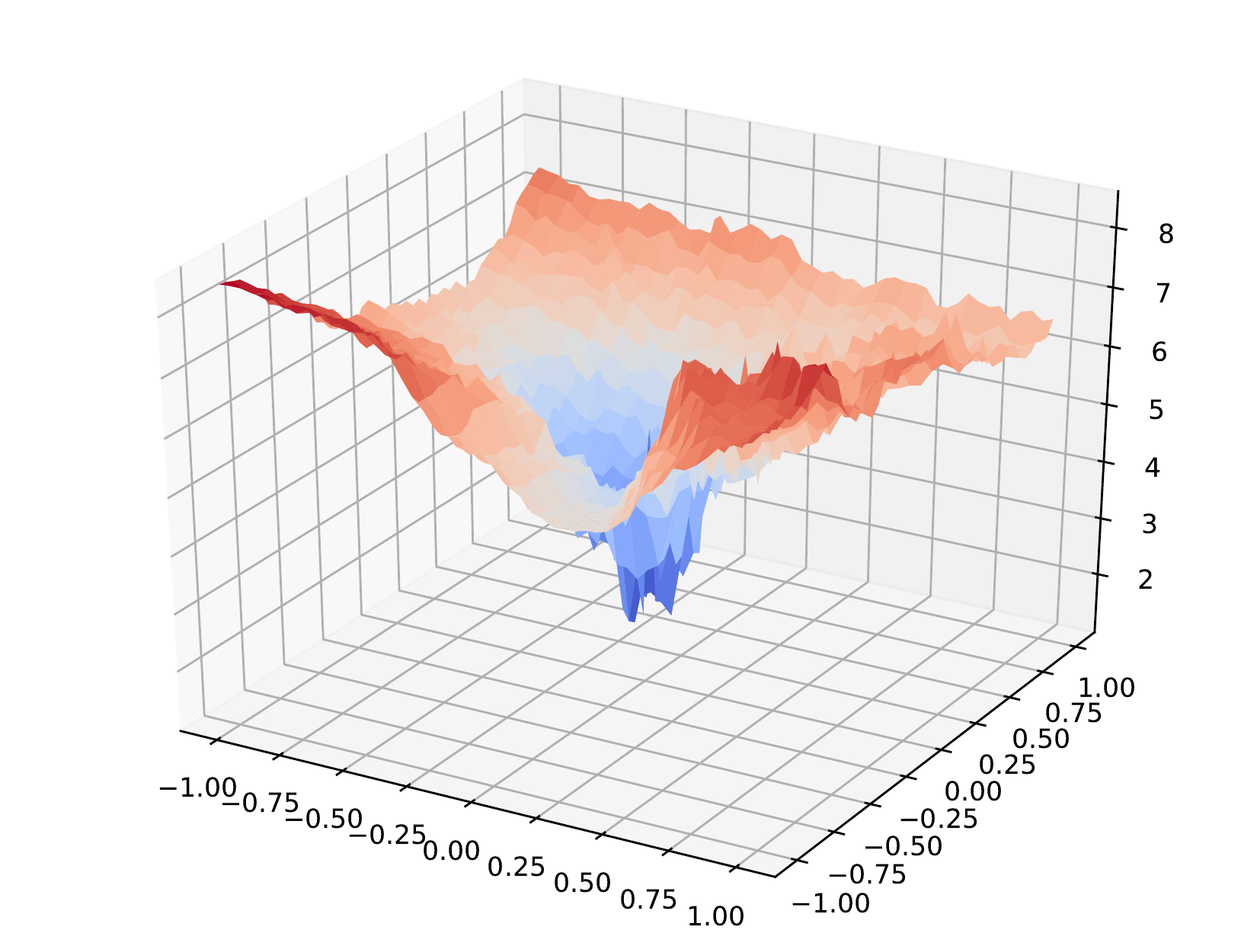}
    \centering
    \caption{BiRealNet-18 + Info-RCP (ours)}
    \label{fig:loss landscape birealnet-18 + info-rcp 3d}
\end{subfigure}
\hfill
\begin{subfigure}[t]{.3\textwidth}
    \centering
    \includegraphics[width=42mm, height=42mm]{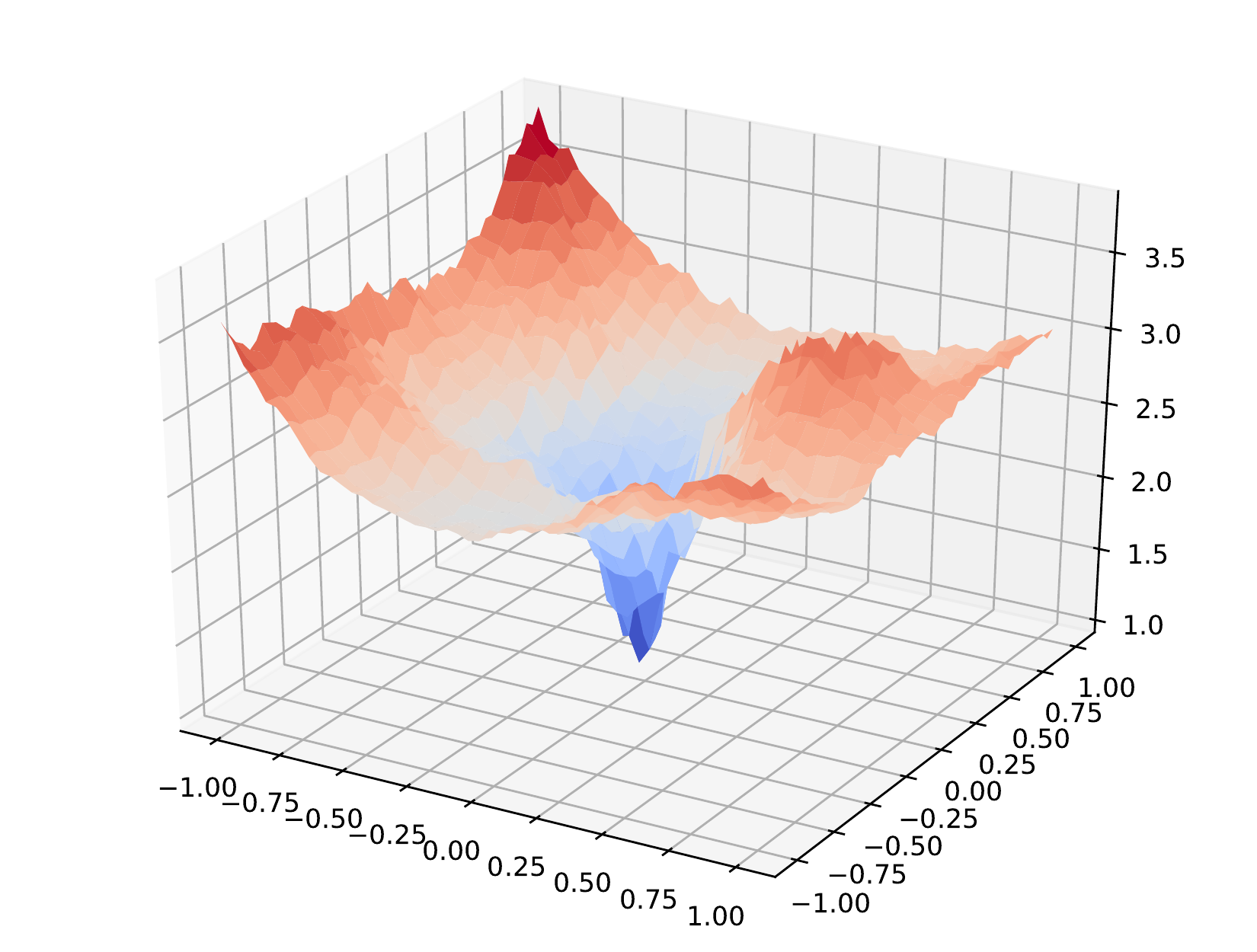}
    \centering
    \caption{BiRealNet-18 + ELM-Attention (ours)}
    \label{fig:loss landscape birealnet-18 + ELM Attention 3d}
\end{subfigure}
\hfill
\begin{subfigure}[t]{.3\textwidth}
    \centering
    \includegraphics[width=42mm, height=42mm]{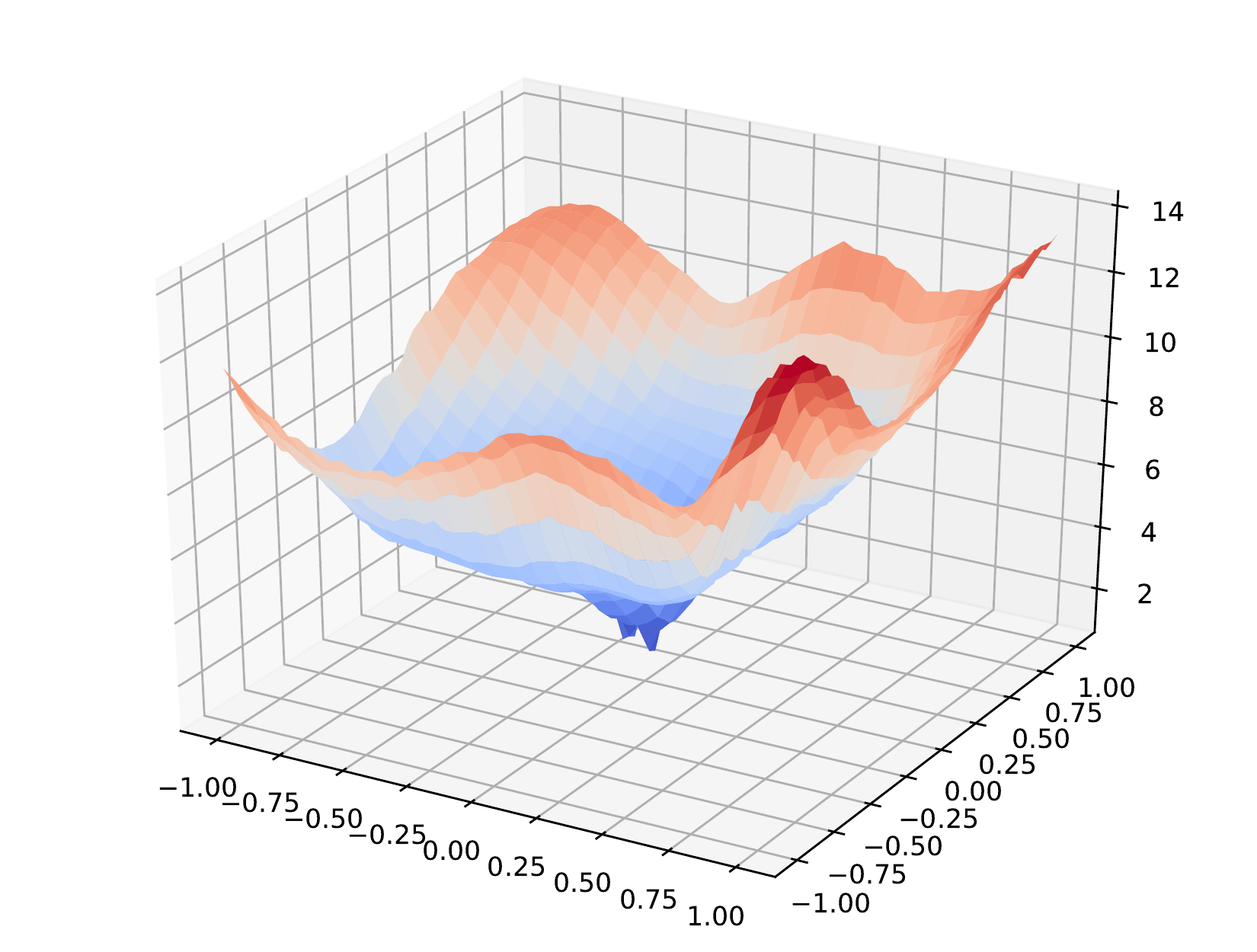}
    \centering
    \caption{BNext-18 (ours)}
    \label{fig:loss landscape bnext 3d}
\end{subfigure}
\hfill
\end{center}
\caption{The loss landscape of popular BNNs (3D view) on the CIFAR10 dataset.
We follow the setting of \cite{li2018visualizing}.
``Info-RCP'' and ``ELM-Attention'' are the module designs in this paper.
Compared with previous design, the BNext-18 shows a smoother loss landscape.
(The corresponding contour lines are shown in Figure~\ref{fig:loss_landscape_visualization} in the main work.)
}
\label{fig:loss_landscape_visualization 3d view}
\end{figure*}
\section{Appendix}
\label{sec:appendix}
In this section, we present more detailed visualization results and an ablation study, which are not listed in the main paper due to limited space.

\subsection{3D Loss Landscape Visualization}
\label{appendix: 3D loss landscape visualization}

We use the loss landscape visualization technique on existing binary architecture designs such as BinaryNet \cite{courbariaux2016binarized}, BiRealNet \cite{liu2018birealnet}, Real2BinaryNet \cite{real2binICLR20}, and their full-precision counterpart ResNet-18 \cite{he2016deep}.
We plotted the corresponding 3D loss landscapes for each network, as shown in Fig.~\ref{fig:loss_landscape_visualization 3d view}.
It also shows the 3D view of the designed modules (Info-RCP and ELM-Attention) individually and our full BNext design (using both modules).
It is easy to see that binarization makes the loss landscape surface really rugged but our design can alleviate this problem of binarization.
As a result, the loss landscape surface of BNext is already close to the previous full precision design.
This explains why BNext can reach a higher accuracy boundary.

\subsection{Counter-Intuitive Overfitting}
\label{appendix: Counter Intuitive Overfitting}
\begin{figure}[]
\captionsetup[subfigure]{justification=centering, font=footnotesize}
\begin{center}
\begin{subfigure}[t]{.45\textwidth}
    \centering
    \includegraphics[width = 0.5\linewidth, scale=0.4]{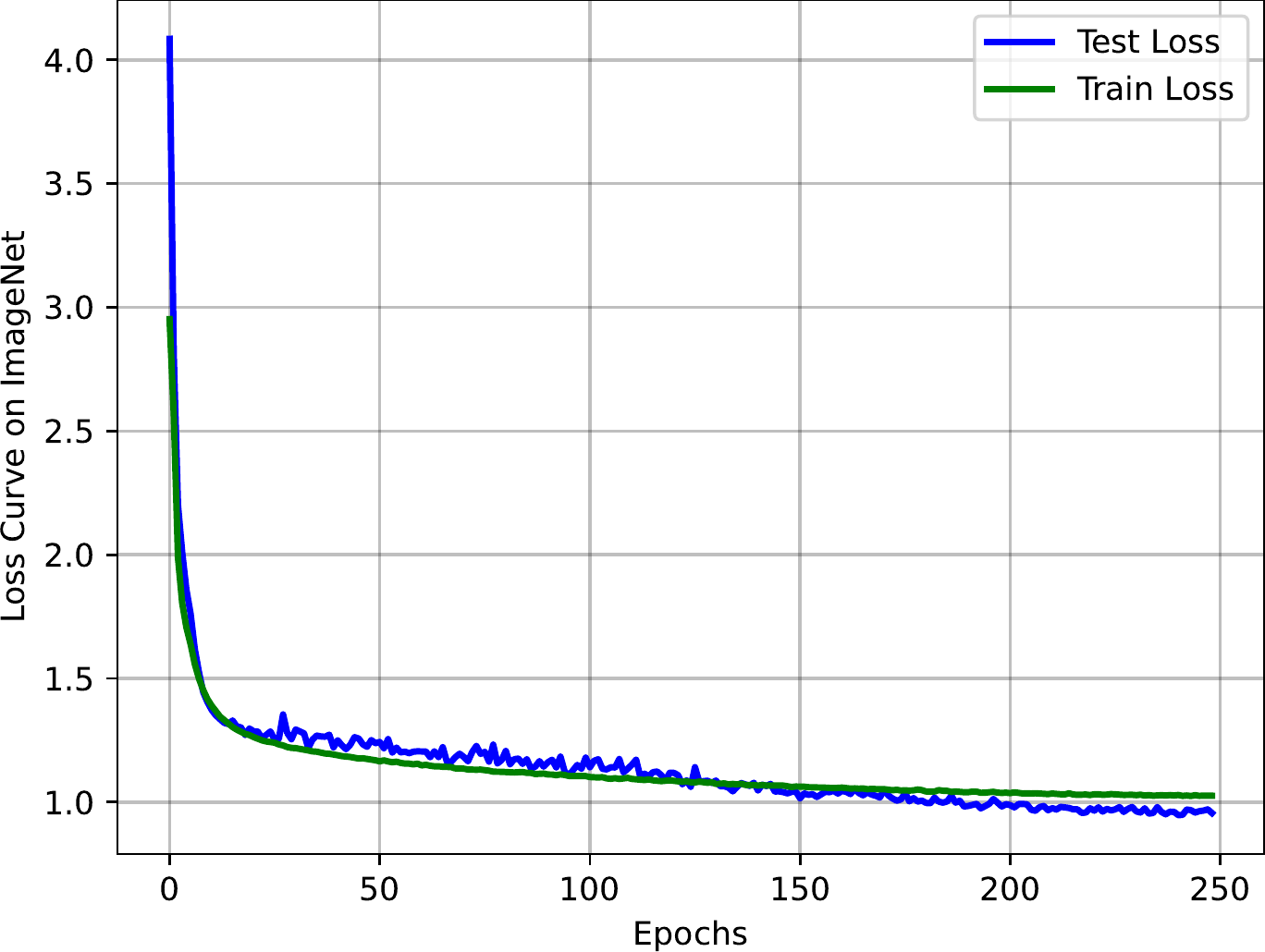}
    \centering
    \caption{BNext-L Loss Curve}
    \label{fig:training loss curve BNext-L  standard KD}
\end{subfigure}
\begin{subfigure}[t]{.45\textwidth}
    \centering
    \includegraphics[width = 0.5\linewidth, scale=0.4]{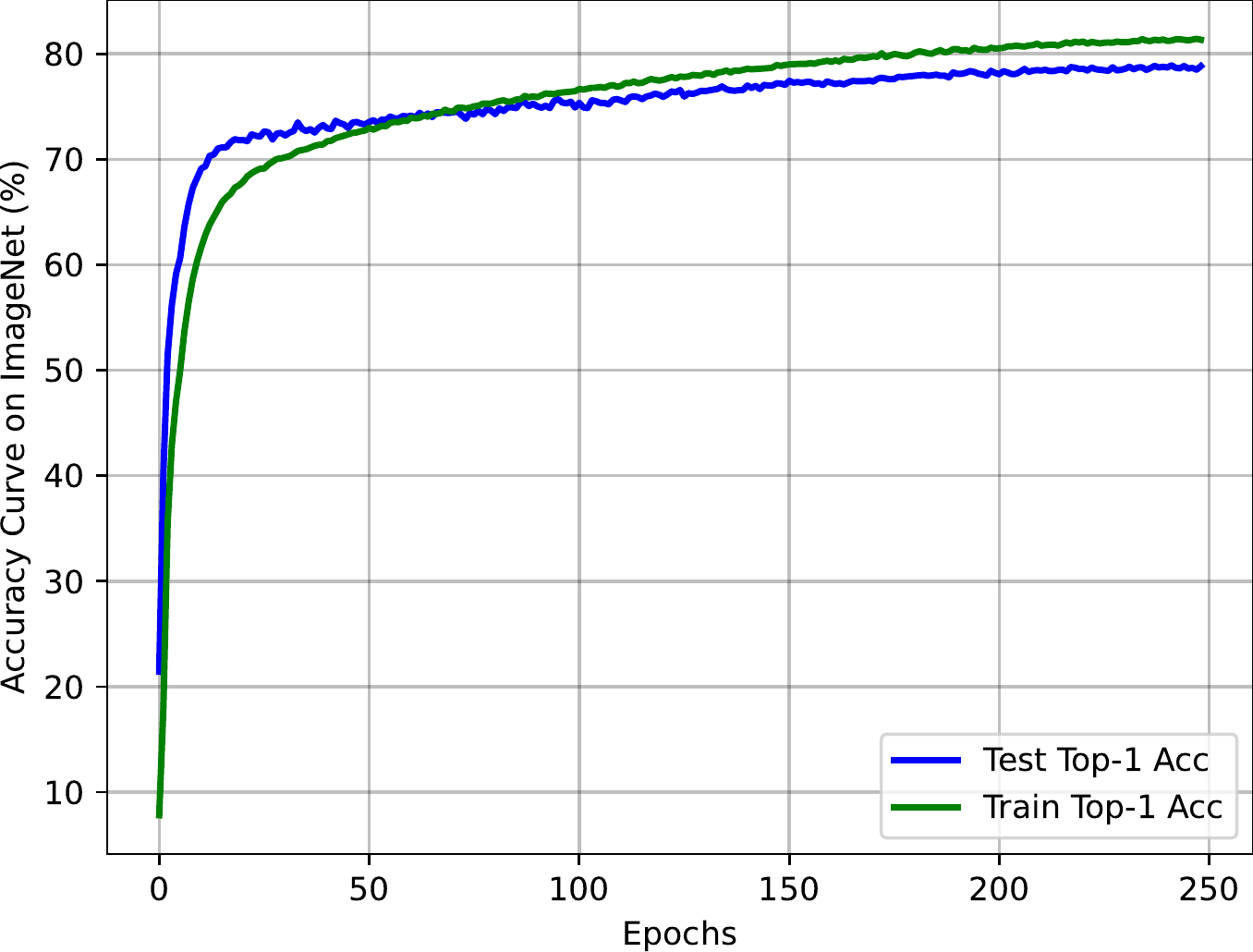}
    \centering
    \caption{BNext-L Accuracy Curve}
    \label{fig:training accuracy curve BNext-L standard KD}
\end{subfigure}
\end{center}
\caption{The optimization procedure of BNext-L model on ImageNet dataset with standard knowledge distillation technique and ConvNext-T as teacher. During the early stage of the optimization, the student always has a lower training loss than testing loss. This indicates a better fitting to training set than testing set.}
\label{fig:overfitting}
\end{figure}
As explained in the main work, knowledge distillation has long been an essential choice for optimizing binary neural networks.
When we move our optimization target from 70\%-level to 80\%-level binary neural network, it is inevitable to seek help from higher accuracy pretrained deep neural networks.
More specifically, we need to decide which model is most suitable as the teacher for the optimization of highly-accuracy BNNs, such as, BNext.
We choose the teacher empirically from a few popular deep neural network families, such as, ResNet, EffcientNet, and ConvNext.
During this process, we observe that the BNext design can easily overfit to strong and highly accurate teachers.
In Table. \ref{tab: Counterintuitive overfitting}, we can see that when a highly accurate full precision model as the teacher leads to a high training accuracy of the student model, but does not generalize well and even performs worse on validation data.
For example, when ConvNext-Base is used as the teacher for BNext-Tiny, the student training accuracy is 3\% higher than using EfficientNet-B0 as the teacher, but the testing accuracy is 1.58\% lower. 

For larger students such as BNext-M/L, this overfitting phenomenon still exists even with increased model capacity.
Take our largest model BNext-L as an example: simply combing standard knowledge distillation with a strong teacher ConvNext-T results in a student with sub-optimal generalization.
As we can see from the optimization procedure in Fig. \ref{fig:overfitting}, the testing loss is always higher than the training loss during the early stage of training procedure. Meanwhile, the performance on the test set is less stable compared to the training curve.
When reaching the same testing accuracy of 78\%, the training accuracy under standard KD is already 80.23\%, which is almost 4\% higher than the final version with the proposed diversified consecutive KD.
After a 256 epochs training, the evaluation result on the test set is 1.67\% lower than the diversified consecutive KD results in Table. \ref{Table:ImageNet SOTA Comparison}.
The overfitting shown in Fig. \ref{fig:overfitting} verifies the need of redesigning the optimization pipeline for BNext.

\begin{table}[]
\centering
\setlength\tabcolsep{2pt}
\footnotesize
\caption{The counter-intuitive overfitting problem observed in the BNext model optimization under different teacher selections.
The BNext-Tiny is used for evaluation.
Each model is optimized for 128 epochs using the standard data augmentation and standard knowledge distillation technique.
``KC" indicates the knowledge-complexity score, Func. \ref{func. knowledge complexity}}
\begin{tabular}{c|cccc}
\hline
Teachers         & KC       &  \thead{Teacher Test \\ Top-1 (\%)}    & \thead{Student Train \\ Top-1 (\%)}                       & \thead{Student Test \\ Top-1 (\%)}                      \\ \hline
None             & 0        &  0     & 76.61                           & 65.02 \\
ResNet-34        & 1.487    &  73.31     & 72.82                           & 71.12 \\
ResNet-50        & 1.539    &  76.13     & 71.59                           & 70.90 \\
ResNet-101       & 1.768    &  77.37     & 72.38                           & 70.76 \\
EfficientNet-B0  & 0.893    &  77.53     & 72.16                           & 71.32 \\
EfficientNet-B2  & 1.103    &  78.06     & 71.91                           & 71.25 \\
EfficientNet-B4  & 1.405    &  79.26     & 73.88                           & 70.63 \\
ConvNext-T       & 1.552    &  82.52     & 73.63                           & 70.63 \\
ConvNext-S       & 1.786    &  83.66     & 74.29                           & 70.43 \\
ConvNext-B       & 2.027    &  84.06     & 75.22                           & 69.74 \\
ResNext101-32x8D & 2.053    &  79.32     & 75.42                           & 69.67                         \\ \hline
\end{tabular}
\label{tab: Counterintuitive overfitting}
\end{table}
\label{appendix: overfitting phenominen}

\subsection{Detailed Teacher Selection for BNext Family Optimization}
\label{appendix: teacher selection for bnext family}
\begin{table}[]
\centering
\setlength\tabcolsep{0.1pt}
\footnotesize
\caption{The detailed teacher settings for our BNext model optimization.}
\begin{tabular}{l|c|c}
\hline
Students & Strong Teacher  & Assistant Teacher Groups                       \\ \hline
BNext-18 & ResNet-34       & None                                            \\
BNext-T  & EfficientNet-B2 & EfficientNet-B0                                 \\
BNext-S  & EfficientNet-B2 & EfficientNet-B0                                 \\
BNext-M  & EfficientNet-B4 & EfficientNet-B0, EfficientNet-B2                \\
BNext-L  & ConvNext-T   & EfficientNet-B0, EfficientNet-B4, ConvNext-T \\ \hline
\end{tabular}%
\label{tab:teacher settings}
\end{table}

To solve the counter intuitive overfitting problems for BNext family optimization, we designed the diversified consecutive KD in Sec. \ref{subsec: knowledge distillation optimization}.
For each BNext design, we build the corresponding knowledge matrix based on a comprehensive consideration of student capacity, teacher knowledge complexity and teacher evaluation performance.
The detailed teacher selections are shown in Table. \ref{tab:teacher settings}.
For BNext-18, we only use the standard KD setting for a fair comparison, which means that only ResNet-34 is utilized during the training procedure.
For BNext-T/S, we apply a gap-aware ensemble strategy for knowledge distillation (Sec. \ref{subsec: knowledge distillation optimization}) to increase the diversity of the teacher knowledge.
Since the capacity of BNext-T/S performance is relatively limited, the EfficientNet-B0/B2 models are already sufficiently strong as teachers.
Consequently, no knowledge boosting strategy is utilized for them.
For BNext-M/L, we rely on higher accuracy teachers like ConvNext-T to explore the accuracy boundary of our BNext design.
As we have observed, simply learning from this kind of high accuracy teacher from scratch under standard KD can suffer from overfitting problem.
Consequently, we use the knowledge-boosting strategy (proposed in Sec. \ref{subsec: Knowledge-Boosting Strategy}) to evolve the supervision information during the training.
In this way, the knowledge discrepancy between student and teacher can be decreased by slowly increasing the knowledge confidence during the training.

\subsection{BNext Model Training Procedure}
\label{appendix: training procedure}
\begin{figure*}[]
\captionsetup[subfigure]{justification=centering, font=footnotesize}
\begin{center}
\begin{subfigure}[t]{.33\textwidth}
    \centering
    \includegraphics[width=48mm, height=38mm]{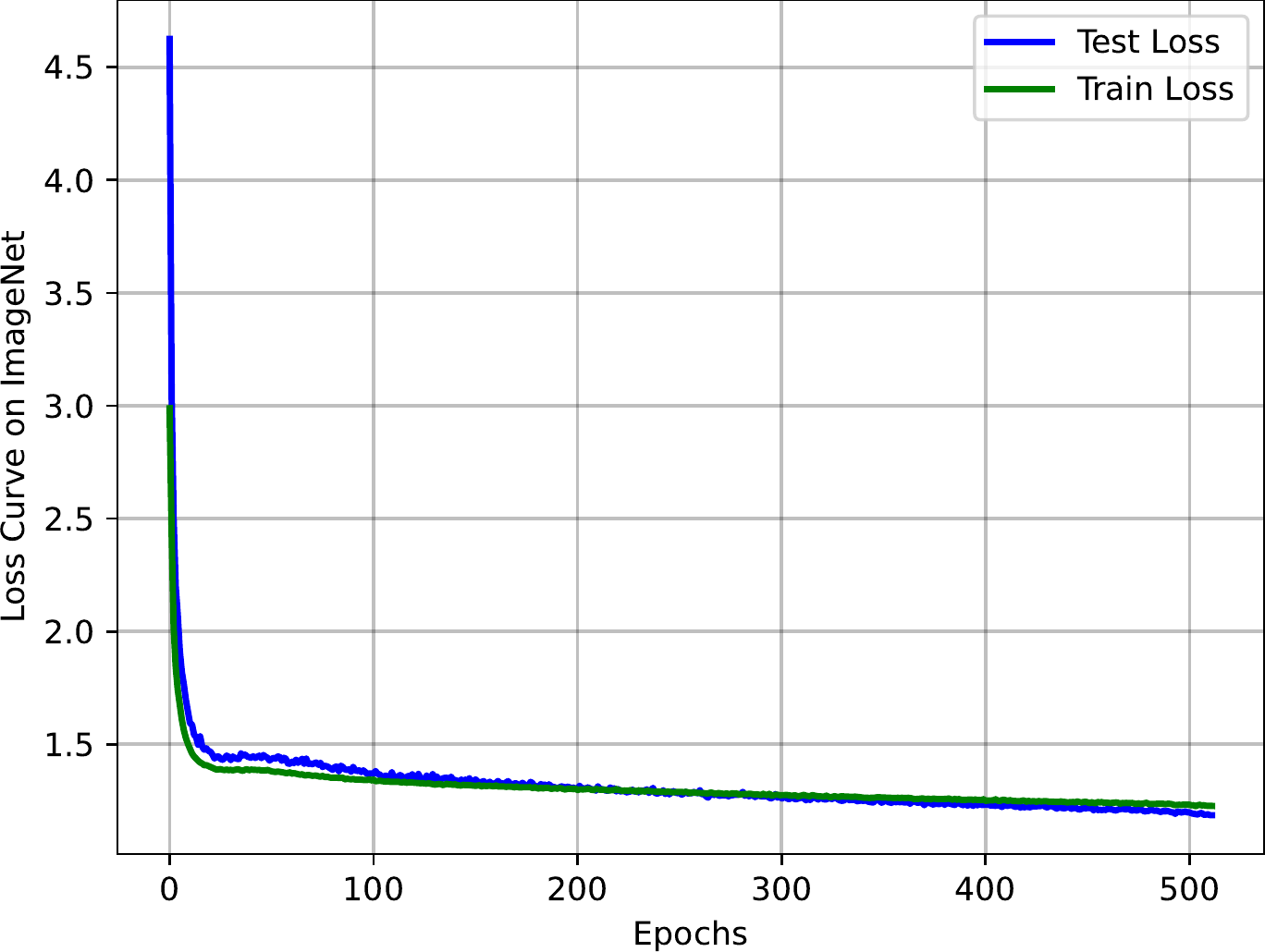}
    \centering
    \caption{BNext-T Loss Curve}
    \label{fig: training loss curve BNext-T}
\end{subfigure}
\hfill
\begin{subfigure}[t]{.33\textwidth}
    \centering
    \includegraphics[width=48mm, height=38mm]{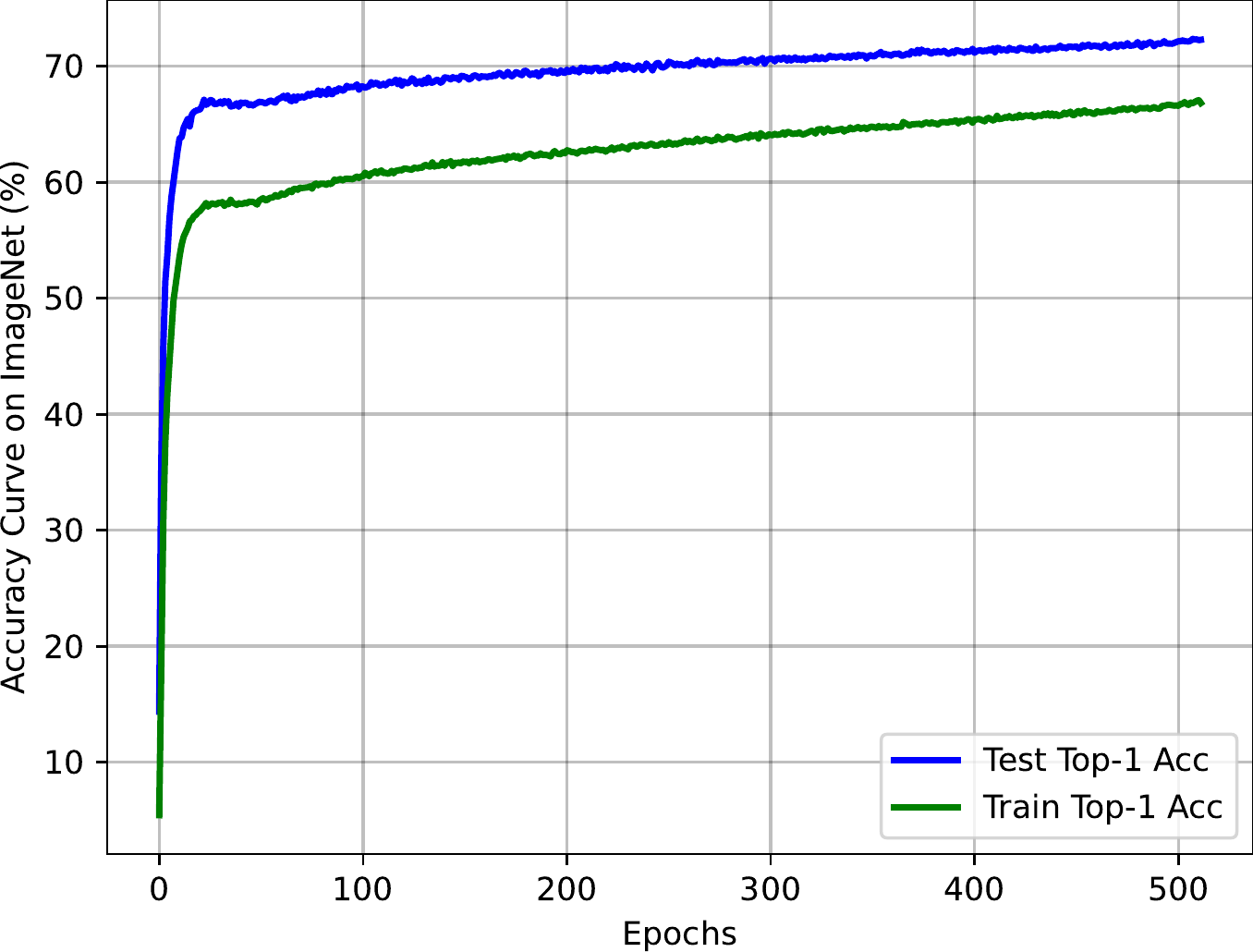}
    \centering
    \caption{BNext-T Accuracy Curve}
    \label{fig:training accuracy curve BNext-T}
\end{subfigure}
\hfill
\begin{subfigure}[t]{.33\textwidth}
    \centering
    \includegraphics[width=48mm, height=38mm]{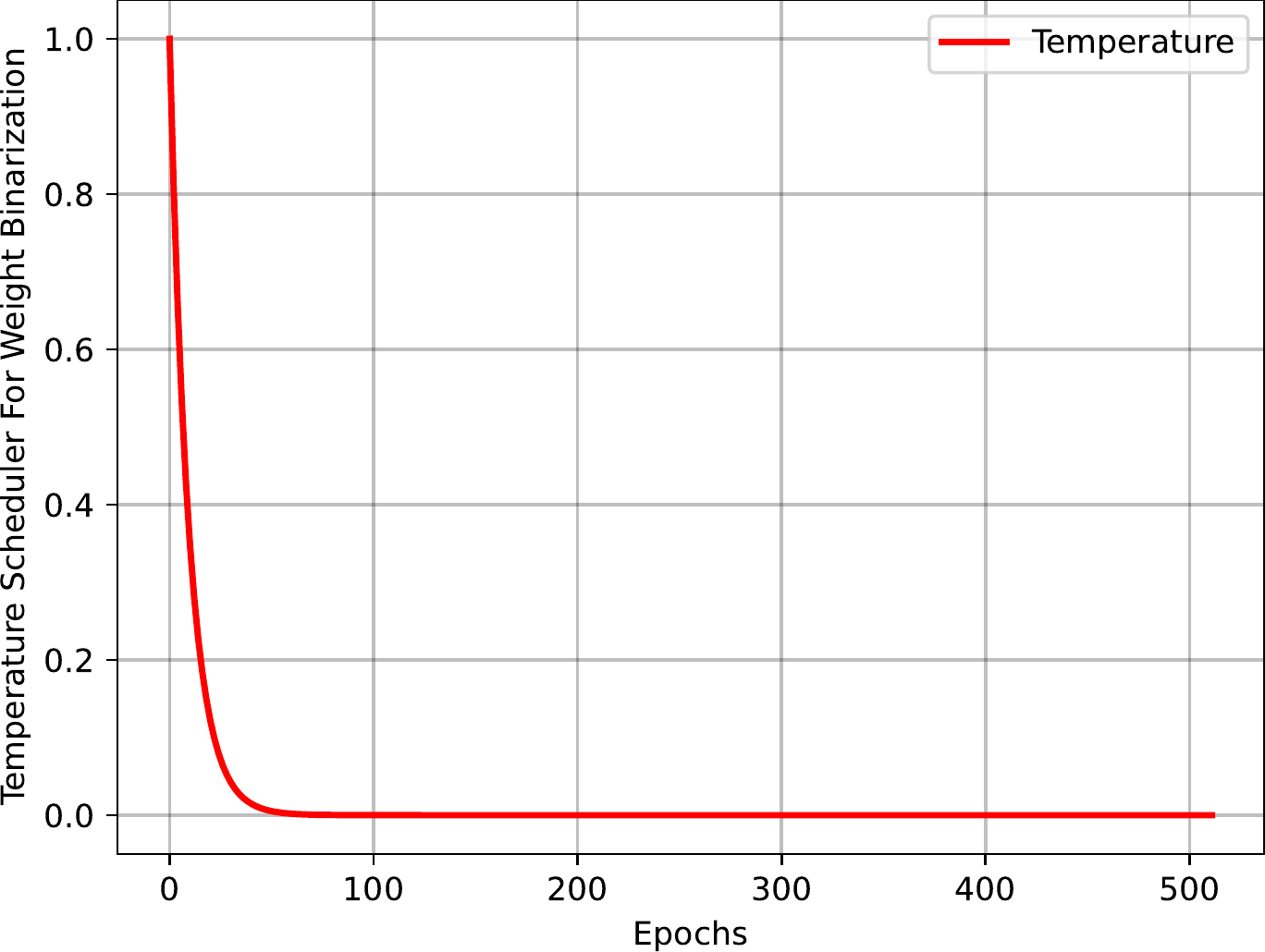}
    \centering
    \caption{BNext-T Temperature Curve}
    \label{fig:training temperature curve BNext-T}
\end{subfigure}
\begin{subfigure}[t]{.33\textwidth}
    \centering
    \includegraphics[width=48mm, height=38mm]{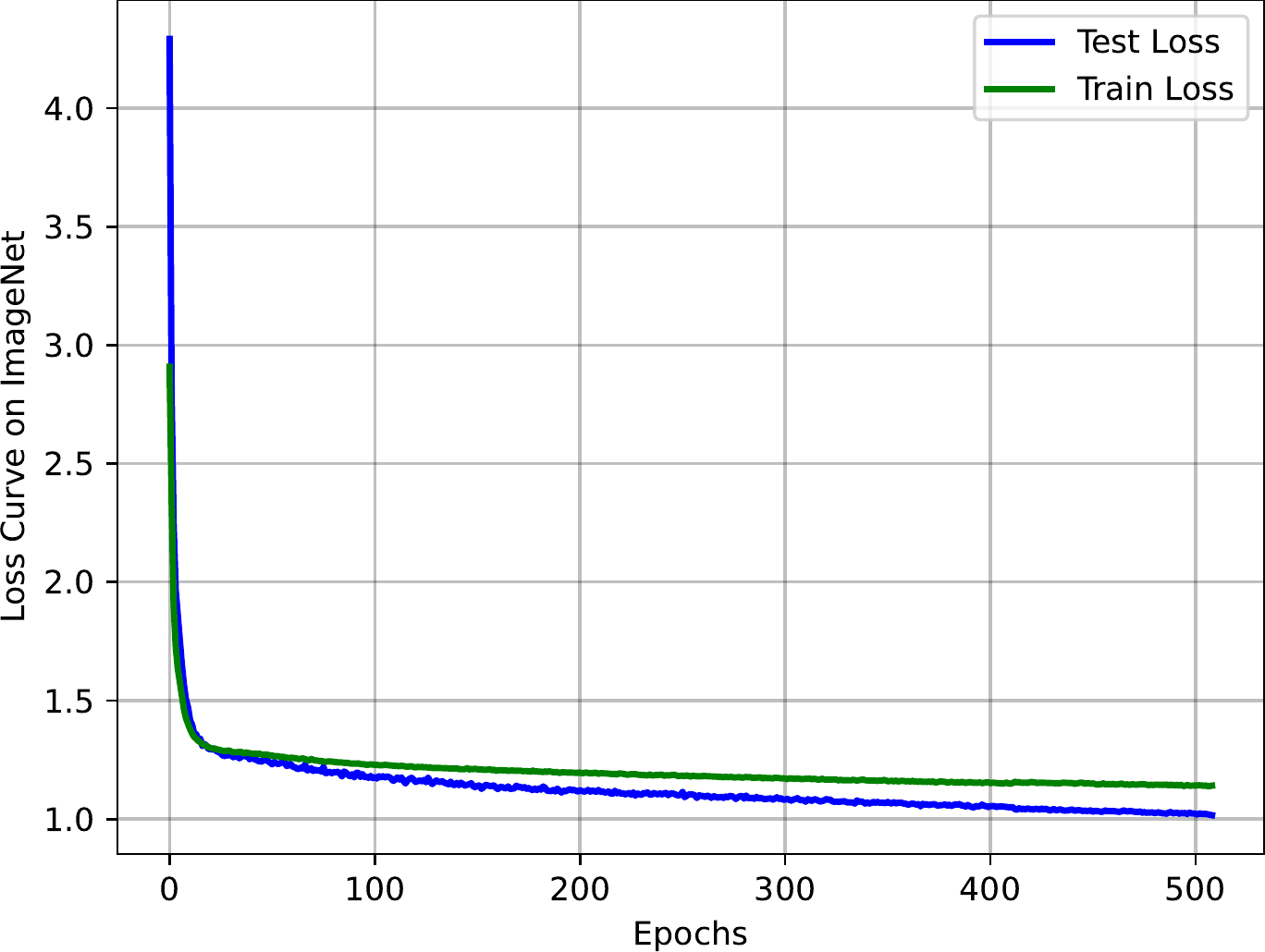}
    \centering
    \caption{BNext-S Loss Curve}
    \label{fig:training loss curve BNext-S}
\end{subfigure}
\hfill
\begin{subfigure}[t]{.33\textwidth}
    \centering
    \includegraphics[width=48mm, height=38mm]{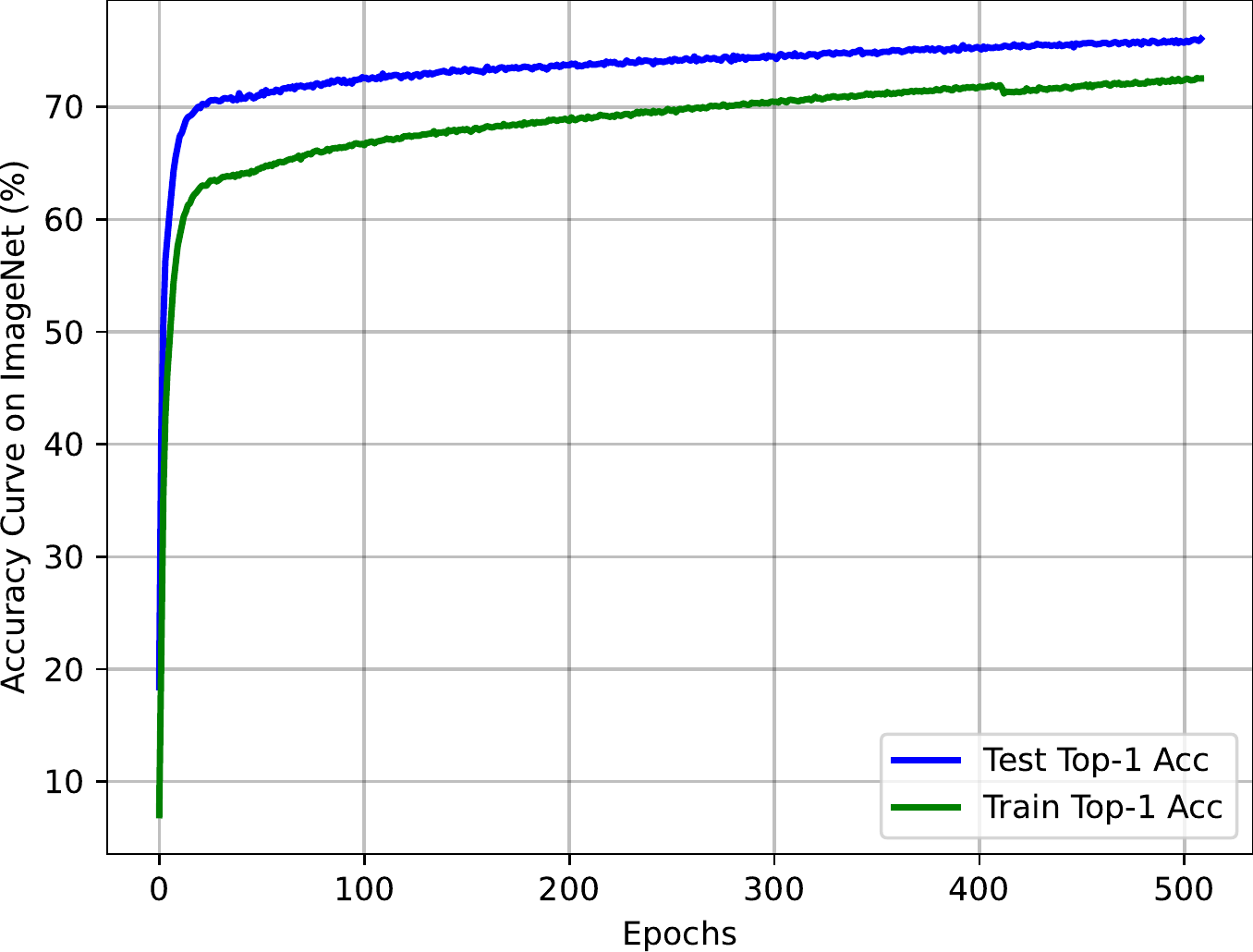}
    \centering
    \caption{BNext-S Accuracy Curve}
    \label{fig:training accuracy curve BNext-S}
\end{subfigure}
\hfill
\begin{subfigure}[t]{.33\textwidth}
    \centering
    \includegraphics[width=48mm, height=38mm]{BNext_Large_Temperature.pdf}
    \centering
    \caption{BNext-S Temperature Curve}
    \label{fig:training temperature curve BNext-S}
\end{subfigure}
\begin{subfigure}[t]{.33\textwidth}
    \centering
    \includegraphics[width=48mm, height=38mm]{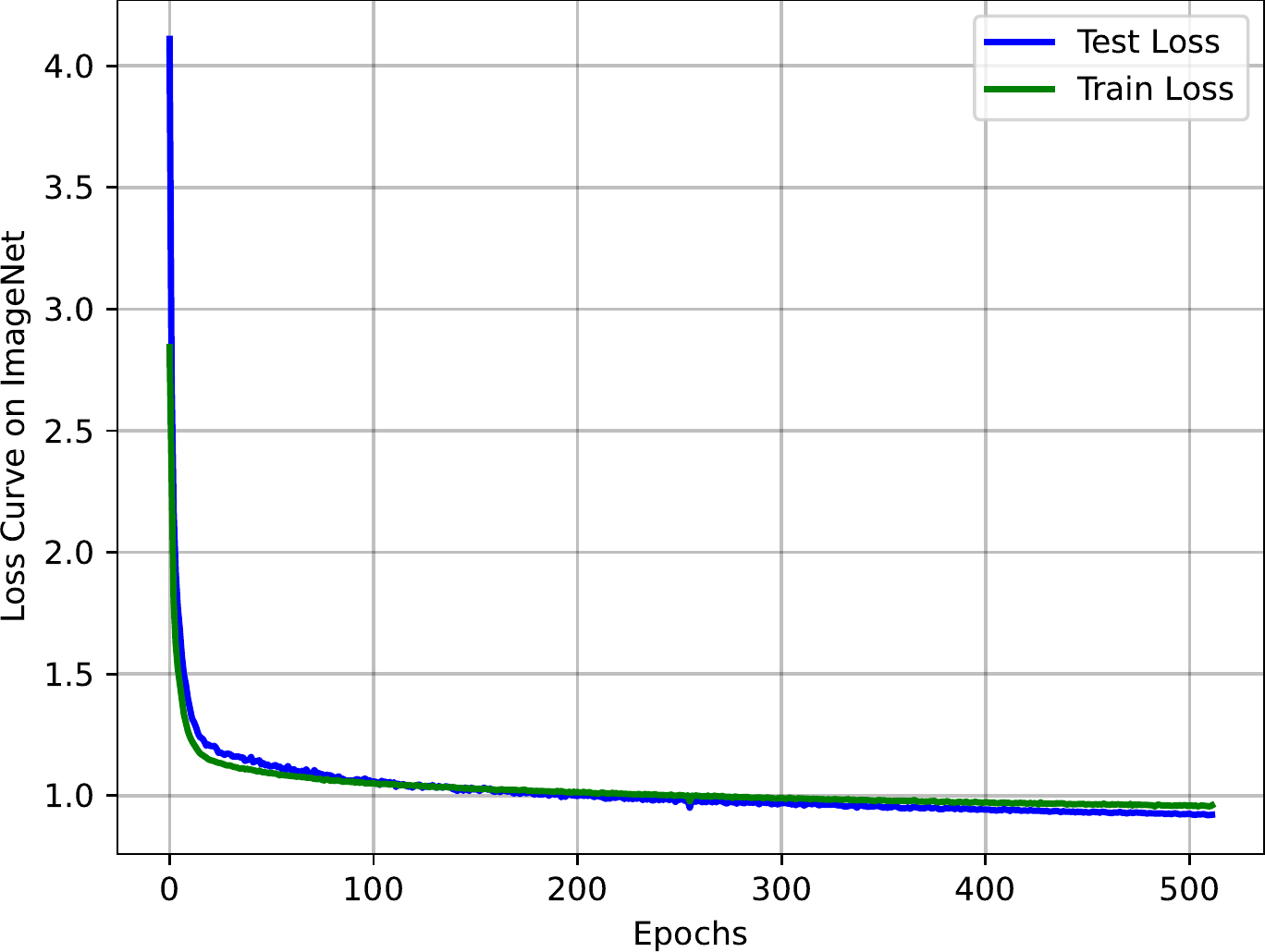}
    \centering
    \caption{BNext-M Loss Curve}
    \label{fig:training loss curve BNext-M}
\end{subfigure}
\hfill
\begin{subfigure}[t]{.33\textwidth}
    \centering
    \includegraphics[width=48mm, height=38mm]{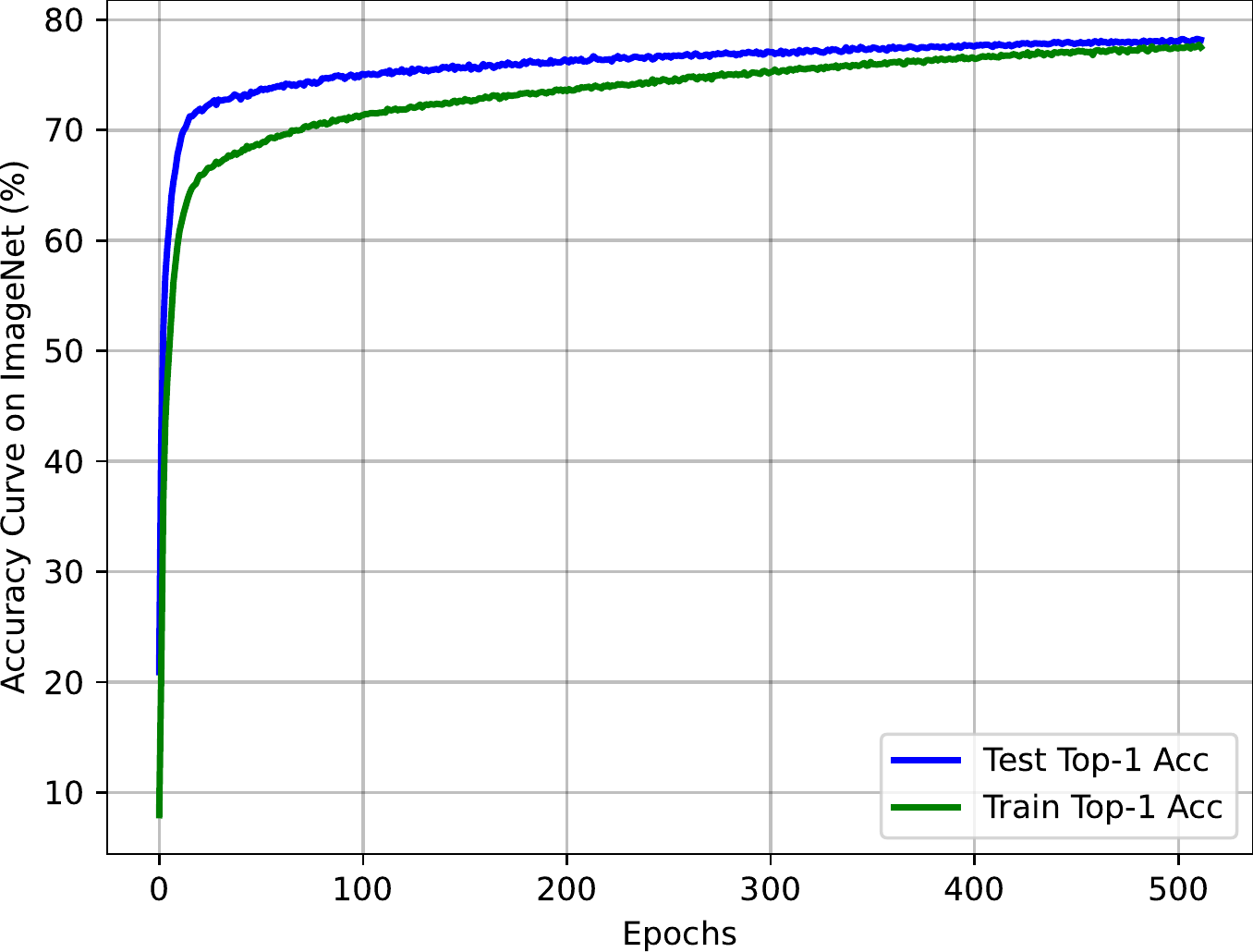}
    \centering
    \caption{BNext-M Accuracy Curve}
    \label{fig:training accuracy curve BNext-M}
\end{subfigure}
\hfill
\begin{subfigure}[t]{.33\textwidth}
    \centering
    \includegraphics[width=48mm, height=38mm]{BNext_Large_Temperature.pdf}
    \centering
    \caption{BNext-M Temperature Curve}
    \label{fig:training temperature curve BNext-M}
\end{subfigure}
\begin{subfigure}[t]{.33\textwidth}
    \centering
    \includegraphics[width=48mm, height=38mm]{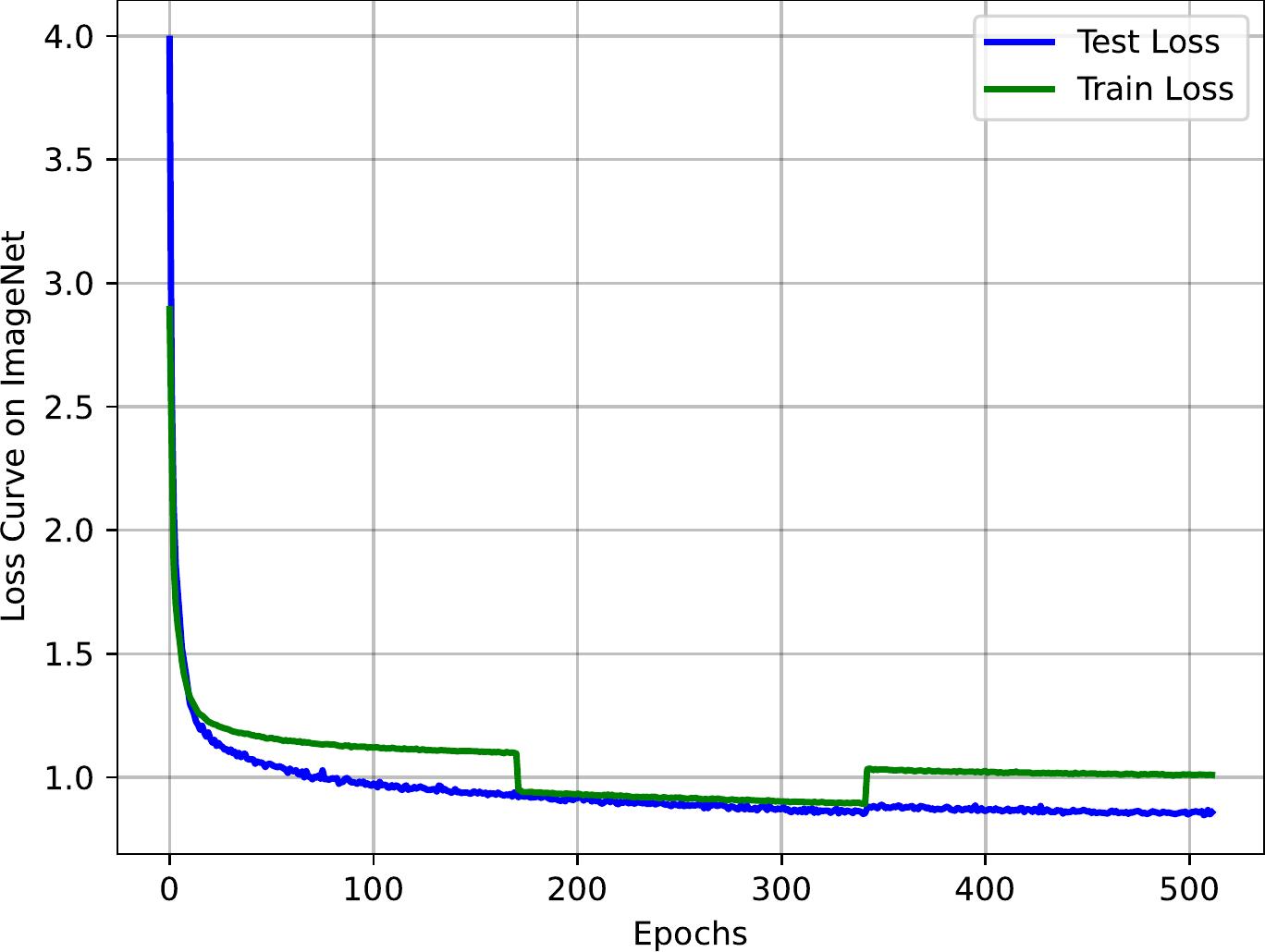}
    \centering
    \caption{BNext-L Loss Curve}
    \label{fig:training loss curve BNext-L}
\end{subfigure}
\hfill
\begin{subfigure}[t]{.33\textwidth}
    \centering
    \includegraphics[width=48mm, height=38mm]{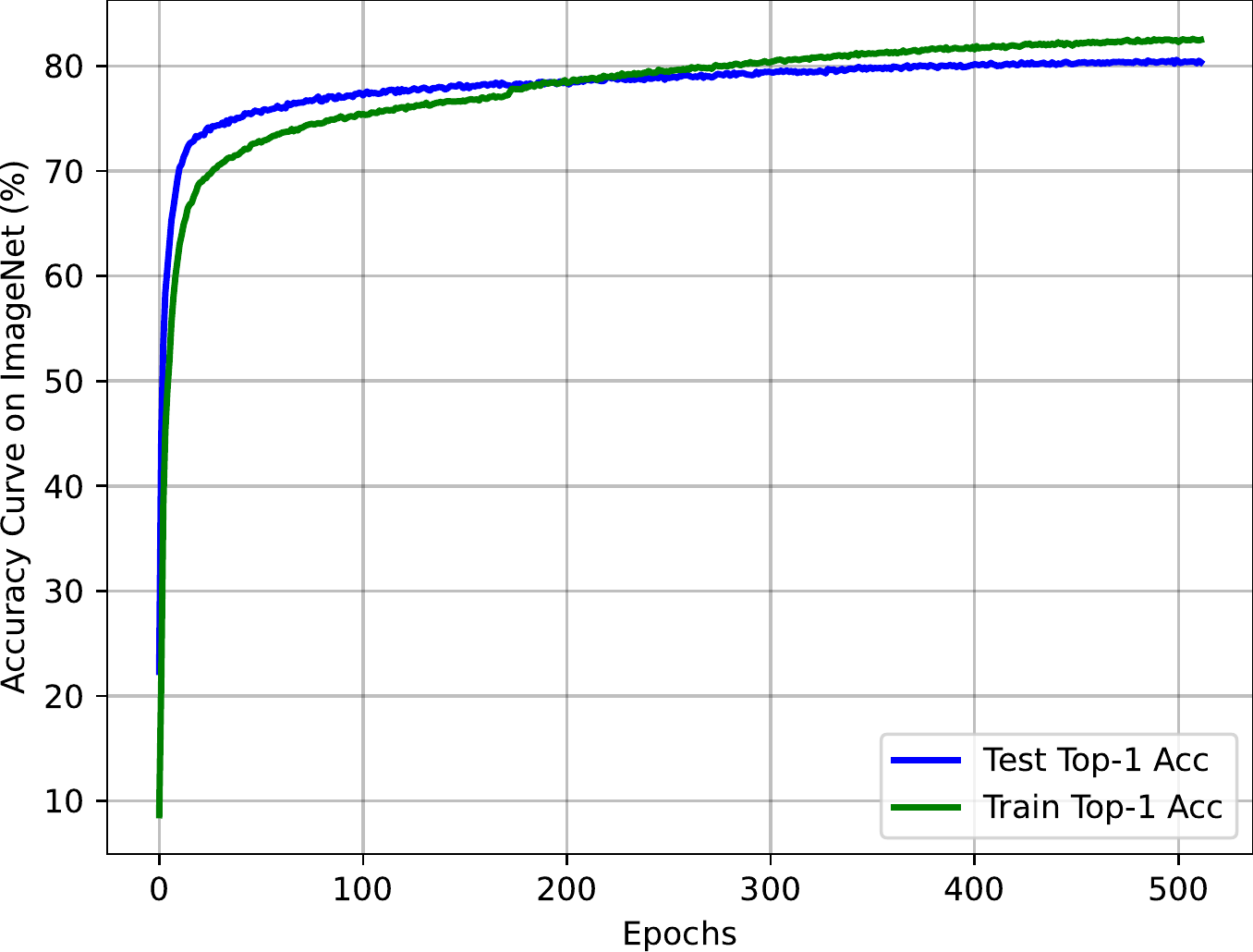}
    \centering
    \caption{BNext-L Accuracy Curve}
    \label{fig:training accuracy curve BNext-L}
\end{subfigure}
\hfill
\begin{subfigure}[t]{.33\textwidth}
    \centering
    \includegraphics[width=48mm, height=38mm]{BNext_Large_Temperature.pdf}
    \centering
    \caption{BNext-L Temperature Curve}
    \label{fig:training temperature curve BNext-L}
\end{subfigure}
\end{center}
\caption{The training procedure of BNext family on ImageNet dataset. The temperature curve indicates the temperature scheduler for the progressive weights binarization.}
\label{fig:bnext-l training procedure}
\end{figure*}

The detailed training procedure of the BNext family (T/S/M/L) are shown in Fig. \ref{fig:bnext-l training procedure}.
Since each BNext design holds different teacher matrix and variant teacher scheduler, we can observe different patterns in the training loss curve and training accuracy curve of each model.
All in all, we can see that each BNext model always has a lower testing loss than training loss during the optimization. If we further compare the testing curves between Fig. \ref{fig:training loss curve BNext-L} and Fig. \ref{fig:training loss curve BNext-L  standard KD}, we can conclude that the proposed diversified consecutive KD helps BNext optimization to generalize better on testing set.

\subsection{BNext Post-Quantization Details}
\label{appendix: post-quantization}

We try to maximize the efficiency of BNext family by further quantizing the input layer, the last layer and the Squeeze-and-Expand branch in each Info-RCP module.
Specifically, we utilize post-training quantization for each BNext model.
For the optimization settings, we use the AdamW optimizer ($\beta_{1} {=} 0.99$, $\beta_{2} {=} 0.999$) to fine-tune the corresponding layers but keep the pretrained binary layers fixed.
The initial learning rate is 1e-7 and each model is fine-tuned for 5 epochs on the ImageNet dataset.
The data augmentation is kept in line with the pre-training phase.
No weight decay is used during this period.
We use asymmetric quantization, which can be mathematically formulated as follows:
\begin{multline}
    X_q^i = \frac{round((X_r^i - \beta)\cdot\alpha)}{\alpha} + \beta, \\
    \Rightarrow \alpha = \frac{2^n-1}{max_{X_r^i}-min_{X_r^i}}, \beta = min_{X_r^i},
\end{multline}
\begin{multline}
    W_q^i = (\frac{round((W_r^{i,Scaled}) - \beta)\cdot\alpha)}{\alpha} + \beta)\cdot Norm(W_r^{i}), \\
    \Rightarrow \alpha = \frac{2^n-1}{max_{W_r^{i,Scaled}}-min_{W_r^{i, Scaled}}}, \beta = min_{W_r^{i, Scaled}},
\end{multline}
where $X^i$ represents the input features and $W^i$ represents weights respectively. The $r$ and $q$ means full precision and quantization representation. The $Norm$ indicates the absolute-mean for each output channel. The $Scaled$ means that the variable is scaled down by its channel-wise absolute-mean.

\subsection{Detailed CIFAR Evaluation Results}

As we have mentioned in the main pages, we train a BNext-18 model on the CIFAR dataset (including CIFAR10 and CIFAR100) to evaluate the generalization of BNext design.
Due to limited space, we only show the averaged results in the main page.
Here, we present the detailed results of 5 runs on CIFAR, as shown in Table. \ref{tab: 5 runs CIFAR10&100}. 
\label{appendix: cifar100 evaluation}

\begin{table}[]
\centering
\caption{5 runs evaluation of BNext-18 on the CIFAR10 and CIFAR100 datasets.}
\label{tab: 5 runs CIFAR10&100}
\setlength\tabcolsep{2.5pt}
\footnotesize
\begin{tabular}{c|cccc}
\hline
Datasets     & \thead{Train \\Top-1 (\%)} & \thead{Train \\Top-5 (\%)} & \thead{Test \\Top-1 (\%)}            & \thead{Test \\Top-5 (\%)}            \\ \hline
CIFAR10-0    & 99.99            & 100              & 93.67                      & 99.86                      \\
CIFAR10-1    & 99.99            & 100              & 93.39                      & 99.79                      \\
CIFAR10-2    & 99.99            & 100              & 93.59                      & 99.83                      \\
CIFAR10-3    & 99.99            & 100              & 93.69                      & 99.84                      \\ 
CIFAR10-4    & 99.99            & 100              & 93.66                      & 99.81                     \\  
Avg  & 99.99            & 100              & 93.60                      & 98.83                     \\
Var  & 0.00             & 0.00             & 0.0152                     & 0.0007                     \\
Std  & 0.00             & 0.00             & 0.1233                     & 0.027                      \\ \hline
CIFAR100-0   & 99.97            & 100              & 72.22                      & 91.23                      \\
CIFAR100-1   & 99.98            & 100              & 72.02                      & 91.75                      \\
CIFAR100-2   & 99.97            & 100              & 72.35                      & 91.61                      \\
CIFAR100-3   & 99.97            & 100              & 71.97                      & 91.51                      \\
CIFAR100-4   & 99.97            & 100              & 72.37                      & 91.64                      \\
Avg & 99.97            & 100              & 72.18                      & 91.55                     \\
Var & 2e-5             & 0.00            & 0.032                       & 0.0379                     \\
Std & 0.0045           & 0.00            & 0.1845                      & 0.1945                     \\ \hline
\end{tabular}%
\end{table}

\subsection{Code}
\label{sec:code}

Within this github link (\url{https://github.com/hpi-xnor/BNext.git}) we provide you with our training code,
so you are able to reproduce our results if desired.

We added all the details needed to reproduce each of our
BNext models depicted in Section \ref{sec:macro architecture design} of our paper in the respective folders:

\begin{itemize}
\item  \textbf{BNext-T:}  \texttt{src/script/BNext-Tiny}
\item  \textbf{BNext-S:}  \texttt{src/script/BNext-Small}
\item  \textbf{BNext-M:} \texttt{src/script/BNext-Middle}
\item  \textbf{BNext-L:}  \texttt{src/script/BNext-Large}
\end{itemize}

The complete code used for each run can be found in the subfolder \texttt{src}
and the exact running command is saved in \texttt{src/script}.
We also added the output logs (\texttt{logs/training.log}).

We use a virtualized environment for PyTorch \cite{paszke2019pytorch} based on Ananconda for our code setup.
The hosts system thus needs support for Python 3.9.13 and a recent NVIDIA CUDA driver (we tested driver version 470.82.01 with CUDA 11.4 before this arxiv submission) for training with GPU.

Note that the ImageNet dataset also needs to be downloaded and prepared manually in the usual manner (using a \emph{train} and \emph{val} folder for the respective split).
The validation images need to be moved into labeled subfolders.

\subsection{Limitations}
\label{appendix: limitations}

The theoretical speed-up is not yet tested on hardware.
Since there is currently no efficient GPU implementation of any BNN, we have started to work on a GPU-accelerated BNext. 
However, as of the arxiv submission of the paper, the results cannot be provided. 
While the contributions of accuracy and theoretical acceleration are sufficient and in line with prior work, we still have good reasons to believe that we can offer our open-source GPU implementation and inference speed verification results in the final version of the paper.

\end{document}